\pdfoutput=1
\documentclass[sigconf]{acmart}
\copyrightyear{2022}
\acmYear{2022}
\setcopyright{acmcopyright}\acmConference[SAC '22]{The 37th ACM/SIGAPP Symposium on Applied Computing}{April 25--29, 2022}{Virtual Event, }
\acmBooktitle{The 37th ACM/SIGAPP Symposium on Applied Computing (SAC '22), April 25--29, 2022, Virtual Event, }
\acmPrice{15.00}
\acmDOI{10.1145/3477314.3507047}
\acmISBN{978-1-4503-8713-2/22/04}

\usepackage{booktabs} % For formal tables
\usepackage{colortbl}
\usepackage{multirow}
\usepackage{xcolor}
\usepackage{subcaption}
\usepackage{rotating}
\usepackage{bm}
\usepackage{amsmath, amsthm, amsfonts}
\usepackage[inline]{enumitem}
\usepackage{bbding}
\usepackage{hyperref}
\usepackage[hyphenbreaks]{breakurl}
\newcommand{\bp}{\ensuremath{\bm{p}}}
\newcommand{\ba}{\ensuremath{\bm{a}}}
\newcommand{\bh}{\ensuremath{\bm{h}}}

\newcommand{\bb}{\ensuremath{\bm{b}}}

\newcommand{\bq}{\ensuremath{\bm{q}}}

\newcommand{\bP}{\ensuremath{\bm{P}}}

\newcommand{\bA}{\ensuremath{\bm{A}}}

\newcommand{\bI}{\ensuremath{\bm{I}}}

\newcommand{\bC}{\ensuremath{\bm{C}}}
\newcommand{\bU}{\ensuremath{\bm{U}}}

\definecolor{colorblue}{rgb}{0.29, 0.59, 0.82}
\definecolor{colorred}{rgb}{0.9, 0.17, 0.31}
\definecolor{colorred}{rgb}{0.9, 0.17, 0.31}
\definecolor{coolgrey}{rgb}{0.55, 0.57, 0.67}
\definecolor{colorwhite}{rgb}{1, 1, 1}

\newcolumntype{"}{@{\hskip\tabcolsep\vrule width 1.5pt\hskip\tabcolsep}}
\makeatletter
\newcommand{\thickhline}{%
    \noalign {\ifnum 0=`}\fi \hrule height 1pt
    \futurelet \reserved@a \@xhline
}

\let\oldTheta\theta
\renewcommand{\theta}[1]{\ensuremath{\oldTheta_{#1}}}

\let\oldSum\sum
\renewcommand{\sum}[2]{\ensuremath{\oldSum_{#1}^{#2}}}
\let\oldProd\prod
\renewcommand{\prod}[2]{\ensuremath{\oldProd_{#1}^{#2}}}

\begin{document}
\title{A Better Loss for Visual-Textual Grounding}
%\titlenote{Produces the permission block, and copyright information}
%\subtitle{}
%\subtitlenote{The full version of the author's guide is available as \texttt{acmart.pdf} document}
  
\renewcommand{\shorttitle}{A Better Loss for Visual-Textual Grounding}

\author{Davide Rigoni}
% \authornote{Dr.~Trovato insisted his name be first.}
\orcid{1234-5678-9012}
\affiliation{%
  \institution{University of Padua \\ Bruno Kessler Foundation}
  \streetaddress{Via Trieste, 63}
  \city{Padua} 
  \state{Italy} 
  \postcode{35121}
}
\email{drigoni@fbk.eu}

\author{Luciano Serafini}
% \authornote{The secretary disavows any knowledge of this author's actions.}
\affiliation{%
  \institution{Bruno Kessler Foundation}
  \streetaddress{Via Sommarive, 18}
  \city{Povo} 
  \state{Italy} 
  \postcode{38123}
}
\email{serafini@fbk.eu}

\author{Alessandro Sperduti}
%\authornote{This author is the one who did all the really hard work.}
\affiliation{%
  \institution{University of Padua}
  \streetaddress{Via Trieste, 63}
  \city{Padua} 
  \state{Italy}
  \postcode{35121}}
\email{sperduti@unipd.it}

%\renewcommand{\shortauthors}{D. Rigoni et al.}

%%%%%%%%% ABSTRAC
\begin{abstract}
Given a textual phrase and an image, the visual grounding problem is the task of locating the content of the image referenced by the sentence.
It is a challenging task that has several real-world applications in human-computer interaction, image-text reference resolution, and video-text reference resolution. 
In the last years, several works have addressed this problem by proposing more and more large and complex models that try to capture visual-textual dependencies better than before. 
These models are typically constituted by two main components that focus on how to learn useful multi-modal features for grounding and how to improve the predicted bounding box of the visual mention, respectively. Finding the right learning balance between these two  sub-tasks is not easy, and the current models are not necessarily optimal with respect to this issue.
In this work, we propose a loss function based on bounding boxes classes probabilities that: (i) improves the bounding boxes selection; (ii) improves the bounding boxes coordinates prediction.
Our model, although using a simple multi-modal feature fusion component, is able to achieve a higher accuracy than state-of-the-art models on two widely adopted datasets, reaching a better learning balance between the two sub-tasks mentioned above.

%Code with all the reproducible experiments are made available online (here there is not the link due to double blind review.)
\end{abstract}

\begin{CCSXML}
<ccs2012>
   <concept>
       <concept_id>10010147.10010178.10010224.10010245.10010251</concept_id>
       <concept_desc>Computing methodologies~Object recognition</concept_desc>
       <concept_significance>500</concept_significance>
       </concept>
   <concept>
       <concept_id>10010147.10010178.10010224.10010245.10010250</concept_id>
       <concept_desc>Computing methodologies~Object detection</concept_desc>
       <concept_significance>500</concept_significance>
       </concept>
 </ccs2012>
\end{CCSXML}

\ccsdesc[500]{Computing methodologies~Object recognition}
\ccsdesc[500]{Computing methodologies~Object detection}

%%%%%%%%% KEYWORDS
\keywords{Computer Vision, Visual Textual Grounding, Semantic Loss}

\maketitle

%%%%%%%%% CHAPTERS
\section{Introduction}
In the last years, the scientific community has devoted much effort in developing deep learning models for computer vision and natural language processing, thanks to the increasing computational resources and the availability of new data.
While deep learning models for computer vision aim to interpret and understand the visual world made by images and videos, deep learning models for natural language processing aim to interpret and understand the human natural language.
In the last decade, these two research areas have made outstanding advancements that have lead to the formulation of more complex problems in which both vision an textual information are required, such as visual-question answering~\cite{antol2015vqa, shih2016look, zhou2015simple}, image retrieval~\cite{mao2014deep, klein2014fisher, DBLP:conf/nips/FromeCSBDRM13, kiros2014unifying}, and visual grounding~\cite{rohrbach2016grounding, chen2018knowledge, zhang2018grounding, gupta2020contrastive}.
Among these, of particular interest is the visual grounding problem, defined as the task of locating the content of the image referenced by a given sentence, a building block for many other real-world applications and more complex tasks.
It is a challenging task, which requires the semantic understanding of the image content and its textual description,  requiring the ability to predict the parts of the image content referred by a specific descriptive sentence. 
It can be formulated as an object detection task followed by a classification task in which, given an input image and sentence, the goal is to return only the detected object(s) in the image that represent(s) the best semantic match with the sentence. 
In the initial phase of research on this problem, many works have followed this formulation, developing the so called two-stage approach models~\cite{rohrbach2016grounding, DBLP:conf/ijcai/YuYXZ0T18}, while more recent works have chosen to address the problem by a one-stage approach model, in which the object detection and the classification problem are solved jointly~\cite{yang2019fast, sadhu2019zero}.

In the two-stage approach, the visual grounding model receives in input a set of proposal bounding boxes previously extracted by an object proposals extractor, such as Edge Boxes~\cite{zitnick2014edge} and Selective Search~\cite{uijlings2013selective}, or by an object detector, such as Faster R-CNN~\cite{ren2015faster}, Single Shot multibox Detector (SSD)~\cite{liu2016ssd}, or YOLO~\cite{redmon2016you, redmon2018yolov3}.
These proposals, jointly with the given input textual sentence describing the content of the image, constitute the visual grounding model input.
Usually, the model embeds the  sentence in an embedding representation that tries to capture its semantic content. 
Then, the model predicts, for each proposal bounding box, a score that represents how much the content of the bounding box is likely to be referred by the sentence.
Often, the two-stage approach models predict new coordinates for the best predicted proposal, in order to adjust the coordinates to better fit the visual content according to the sentence semantic information.

In the one-stage approach, the visual grounding model receives in input only an image and a textual sentence. 
Then the model learns how to extract and fuse all the visual and textual information in order to predict the best bounding box in output, according to the input sentence.
Even if this seems to be the best approach in order to reach the best results, due to the small number of assumptions made by the model, it raises some major issues:
\begin{enumerate*}[label=(\it\roman*)]
    \item not all the visual grounding datasets are suitable for training an object detector, due to lack of images and/or because they are not densely annotated;
    \item the model requires an high number of parameters, and because of that
    \item the training requires significant computing resources.
\end{enumerate*} 

In the literature, there are many works adopting increasingly improved object proposals, and increasingly complex architectures than before in order to capture the visual and textual information. 
These models are typically constituted by two main components that focus on how to learn useful multi-modal features for grounding, and how to improve the predicted bounding box of the visual mention, respectively. Finding the right learning balance between these two  sub-tasks is not easy, and the current models are not necessarily optimal with respect to this issue.
In this work, we  propose a model that, although using a simple multi-modal feature fusion component, is able to reach a higher accuracy than state-of-the-art models thanks to the adoption of a more effective loss function that reaches a better learning balance between the two sub-tasks mentioned above.

Our main contributions can be summarized as:
\begin{enumerate*}[label=(\roman*)]
    \item we present a new loss for visual proposals which considers also the object proposals semantic information,  differently from the works in the literature which just consider  their shapes and spatial positions in the image;
    \item we are the first to adopt the \textit{Complete Intersection over Union}~\cite{zheng2020enhancing} loss for the visual grounding task;
    \item we introduce a new regression loss on the proposal bounding boxes coordinates which is applied to a subset of all the proposals, selected by considering the object proposals semantic information.
    This loss differs from the one used by the approaches in the literature, which only considers the proposal with the largest overlap with the ground truth. 
    \item we experimentally show that the proposed losses improve the performance of state-of-the-art models.
\end{enumerate*}

\section{Related Works}
\label{sec:sota}
In this section, we report the important related works developed within three areas, namely, Referring Expression Grounding,  Visual-Textual-Knowledge Entity Linking (VTKEL), and Image Retrieval.

\paragraph{Referring Expression Grounding} 
It is also known as phrase localization or visual grounding.
It aims to localize the corresponding objects described by a human natural language phrase in an image.
It is common to extract visual and language features independently and fuse them before the prediction.
Some works apply a multi layer perceptron (MLP)~\cite{chen2017query, chen2018knowledge}, cosine similarity~\cite{engilberge2018deep} and element-wise multiplication.
Other works apply more complex strategies such as Canonical Correlation Analysis (CCA)~\cite{DBLP:conf/iccv/PlummerMCHL17, plummer2015flickr30k}, Multimodal Compact Bilinear (MCB)~\cite{DBLP:conf/emnlp/FukuiPYRDR16}, graph structures~\cite{bajaj2019g3raphground} and attention methods~\cite{nguyen2018improved}.
Instead of focusing on the fusion component, \cite{DBLP:conf/ijcai/YuYXZ0T18} proposes a visual grounding model with diverse and discriminative proposals that can achieve good performance without using a complex multi-modal fusion operator. The approach presented in~\cite{akbari2019multi} predicts the image content location referred by the input phrase using an heatmap, applying a multi-level multi-modal attention mechanism, instead of relying on the standard bounding box. 
Some works focus on the weakly supervised referring expression grounding setting, in which it is available only the information about the image contents and there is no mapping among the input textual sentences and these visual locations.
Given a set of bounding boxes proposals and a textual sentence in input, \cite{rohrbach2016grounding} introduces a model which learns to ground by reconstructing the given textual sentence adopting a soft attention mechanism. This approach is extended in~\cite{chen2018knowledge} by the introduction of a novel Knowledge Aided Consistency Network
(KAC Net) which is optimized by reconstructing the input query and proposal’s information. A different approach is developed in~\cite{xiao2017weakly}, where an end-to-end model learns to localize arbitrary linguistic phrases in the form of spatial attention masks, using two types of carefully designed loss functions. A Variational Context model,  based on the variational Bayesian method, is adopted by
\cite{zhang2018grounding} to exploit the reciprocal relation between the referent and context. The Multi-scale Anchored Transformer Network (MATN)~\cite{zhao2018weakly} hinges on the concept of anchors, i.e. it uses region proposals as localization anchors, learning a multi-scale correspondence network to continuously search for sentences referring to the anchors. The work presented in~\cite{gupta2020contrastive} shows that textual sentence grounding can be learned by optimizing word-region attention to maximize a lower bound on mutual information between images and caption words. 
\cite{liu2019improving} uses a cross-modal attention-guided erasing approach, where it discards the most dominant information from either textual or visual domains to generate difficult training samples online in order to drive the model to discover complementary textual-visual correspondences.
\cite{DBLP:conf/cvpr/DengWWHLT18} provides an accumulated attention (A-ATT) mechanism to ground the natural language
query into the image using a query attention, an image attention and an objects attention.
%\textcolor{red}{Problema! Spesso ci si riferisce a Visual Grounding/Referring Expression Grounding per indicare due task simili. Il primo task e' come il nostro, data una frase intera occorre trovare tutti i riferimenti nell'immagine che sono nominati nel testo, mentre il secondo vuole trovare solo quella principale! ES: "the cat near the towel". Nel primo caso si cerca sia "cat" che "towel", mentre nel secondo caso solo "cat". Purtroppo non c'e' una distinziona nel nome del task e ne segue che si fa confusione. Infatti un riferimento mancante dei reviewers si riferiva proprio ad un lavoro che svolge il secondo task.}

\paragraph{Visual-Textual-Knowledge Entity Linking}
The VTKEL task~\cite{dost2020vtkel, dost2020jointly, dost2020visual} introduces a more complex task than the referring expression task, in which an artificial agent needs to jointly recognize the entities shown in the image and mentioned in the text, and to link them to its prior background knowledge.
The solution to the VTKEL problem could lead to major scientific advancement towards a better understanding of semantic information contained in the image and textual sentence, respectively.
In fact, the knowledge graph allows to introduce semantic reasoning on the information contained in both the image and the textual sentence, which could lead to innovative solutions for the weakly supervised referring expression problem and for the partially annotated dataset problem.

\paragraph{Image Retrieval}
The standard text-base image retrieval systems, given a textual sentence in input, from a set of images select the  one that best matches the textual input.
In particular, the best images are returned according to some metric learned through a recurrent neural network~\cite{mao2014deep}, correlation analysis~\cite{klein2014fisher} and other methods~\cite{DBLP:conf/nips/FromeCSBDRM13, kiros2014unifying}.
\section{Background}
\label{sec:background}
In order to explain our work, we use the following notation:
lower case symbols for scalars and indexes, e.g. $n$;
italics upper case symbols for sets, e.g. $A$;
upper case symbols for textual sentences, e.g. S;
bold lower case symbols for vectors, e.g. $\ba$;
bold upper case symbols for matrices and tensors, e.g. $\bA$;
the position within a tensor or vector is indicated with numeric subscripts, e.g. $\bA_{ij}$ with $i,j \in \mathbb{N}^+$;
calligraphic symbols for domains, e.g. $\mathcal{Q}$. 

In our work we adopt the \textit{Complete IoU (CIoU)}~\cite{zheng2020enhancing} loss to perform the bounding boxes coordinates regression, that is based on the \textit{Intersection over Union (IoU)} metric.
Given a pair of bounding box coordinates $(\bb_i, \bb_j)$, the \textit{Intersection over Union}, also known as Jaccard index, is an evaluation metric used mainly in object detection tasks, which aims to evaluate how much the two bounding boxes refer to the same content in the image.
It is defined as:
\begin{align}
    IoU(\bb_i, \bb_j) = \frac{|\bb_i \cap \bb_j|}{|\bb_i \cup \bb_j|},
\end{align}
where $|\bb_i \cap \bb_j|$ is the area of the box obtained by the intersection of  boxes $\bb_i$ and  $\bb_j$, while $|\bb_i \cup \bb_j|$ is the area of the box obtained by the union of boxes $\bb_i$ and  $\bb_j$.
It is invariant to the bounding boxes sizes, and it returns values that are strictly contained in the interval $[0, 1]\subset \mathbb{R}$, where $1$ means that the two bounding boxes refer to the same image area, while a score of $0$ means that the two bounding boxes do not overlap at all.
The fact that two bounding boxes that do not overlap have \textit{IoU} score equal to $0$, is the major issue of this metric: the zero value does not represent how much the two bounding boxes are far from each other.
For this reason, in its standard definition, the \textit{IoU} function is mainly used as an evaluation metric rather than as a component of a loss function for learning.

%In order to solve the issue of \textit{IoU} when considering it as a loss function, several alternative formulations were suggested in the literature, e.g.
%\cite{rezatofighi2019generalized} proposed the \textit{Generalized IoU (GIoU)} loss, \cite{zheng2020distance} proposed the \textit{Distance IoU (DIoU)} loss, while only recently \cite{zheng2020enhancing} proposed the \textit{Complete IoU (CIoU)} loss, which has shown promising results and faster convergence than \textit{GIoU} and \textit{DIoU}.
In order to solve the issue of \textit{IoU} when considering it as a loss function, \cite{zheng2020enhancing} proposed the \textit{Complete IoU} loss that is defined as:
\begin{align}
\mathcal{L}_{CIoU}(\bb_i, \bb_j)
&=S\left(\bb_i, \bb_j\right)+D\left(\bb_i, \bb_j\right)+V\left(\bb_i, \bb_j\right)\\
S\left(\bb_i, \bb_j\right) &= 1 - IoU(\bb_i, \bb_j); \\
D\left(\bb_i, \bb_j\right) &= \frac{\rho\left(\boldsymbol{p_i}, \boldsymbol{p}_j\right)^{2}}{c^{2}}; \\
\label{eq:5}
V\left(\bb_i, \bb_j\right) &= \alpha \frac{4}{\pi^2} \left(\arctan\frac{wt_{j}}{ht_{j}} - \arctan\frac{wt_{i}}{ht_{i}}\right)
\end{align}
where $\bb_i$ and $\bb_j$ are two bounding boxes, $\boldsymbol{p_i}$ and $\boldsymbol{p}_j$ are their central points, $IoU(\bb_i, \bb_j)$ is the standard \textit{IoU}, $\rho$ is the euclidean distance between the given points, $c$ is the diagonal length of the \textit{convex hull} of the two bounding boxes, 
$\alpha$ is a trade-off parameter, $wt_i$ and $ht_i$ are the width and the height of the bounding box $\bb_i$, respectively.
Differently from the standard \textit{IoU}, the \textit{Complete IoU} is formulated in such a way to return meaningful values, leveraging the bounding boxes geometric shapes, even when two bounding boxes are not overlapped.

\section{Problem Definition}
Visual grounding is the general task of locating the components of a structured description in an image. In order to solve this task, first, it is necessary to recognize all the objects in the image and the components in the text, while after, the model needs to find the correct alignment among the nouns and the objects.
Each detected object in the image is usually represented by a rectangle called bounding box, while each noun phrase detected in the text is usually called query. The bounding box is determined by its position in the image and by its dimension, while the query is determined by the position of the first character and the position of the last character in the input text.

Formally, given in input an image $\bI$ and a sentence S describing some of the objects represented in $\bI$, the task consists in learning a map $\gamma$ from the set $Q$ of noun phrases contained in S to a set of bounding boxes $B$ defined on $\bI$, i.e. $\gamma: \mathcal{I}\times \mathcal{S}\rightarrow 2^{\mathcal{Q}\times \mathcal{B}}$, where $\mathcal{I}$ is the domain of images, $\mathcal{S}$ is the domain of sentences, $\mathcal{Q}$ is the noun phrases domain, $\mathcal{B}$ is the domain of bounding boxes which can be defined on $\mathcal{I}$, and $2^{\mathcal{Q}\times \mathcal{B}}$ is the power set of the Cartesian product between $\mathcal{Q}$ and $\mathcal{B}$.
So, given an image $\bI$ containing $e$ objects identified via the set of bounding boxes $B_{\bI}=\{\bb_i\}_{i=1}^e$, where $\bb_i \in \mathbb{R}^4$ is the vector of coordinates 
identifying a bounding box in $\bI$,  and a sentence S containing $m$ noun phrases gathered in the set 
$Q_{\mbox{S}}=\{ \bq_j \}_{j=1}^m$,  
where $\bq_j \in \mathbb{N}^2$ is a vector containing the initial and final character positions in the sentence S, $\gamma(\bI,\mbox{S})$ returns a subset $\Gamma \subseteq Q_{S}\times B_{\bI}$ %of couples $\{(\bq, \bb)| \bq\in Q_{\mbox{S}},\ \bb \in B_{\bI}\}$ 
where each couple $(\bq, \bb) \in \Gamma$ associates the noun phrase $\bq$ to the bounding box $\bb$. Please, notice that the same  noun phrase can be associated to several different bounding boxes, as well as the same bounding box can be associated to many different noun phrases. Following the current literature, in this paper we assume that each noun phrase is associated to one and only one bounding box. 
A bounding box, however, can identify more objects, e.g. several persons in the case the noun phrase is ``people''. 
A training set of $n$ examples is defined as \mbox{$D = \{(\bI_i, \mbox{S}_i, \Gamma_i^{gt})\}_{i=1}^n$}, where  %$\{(\bq_{j_i}^{gt}, \bb_{j_i}^{gt})\}_{j_i}$ 
$\Gamma_i^{gt}$ is the set of ground truth associations for example $i$. % and query $j$, i.e. $\gamma(\bI_i, \mbox{S}_i) = \{(\bq_{j_i}^{gt}, \bb_{j_i}^{gt})\}_{j_i}$. 
%In the following, we denote with $\Gamma_i^{gt}$
%, where in addition to an image $I$ and a sentence $S$, only a set $B^{gt}$ of ground truth bounding boxes is given.

%In our work we adopt the \textit{Complete IoU (CIoU)}~\cite{zheng2020enhancing} loss to perform the bounding boxes coordinates regression. We refer the reader to the Background section of the Supplementary Material for more details on the standard IoU metric and the CIoU loss.

%In our work we adopt the \textit{Complete IoU (CIoU)}~\cite{zheng2020enhancing} loss to perform the bounding boxes coordinates regression, that is strictly related to the \textit{IoU} metric.

\section{Our Proposal}
\label{sec:model}
% include image of the model
%\begin{picture}(100,100)
%\put(0,0){\includegraphics{img/model.png}}
%\put(10,10){hello}
%\end{picture}
\begin{figure*}[th]
\begin{center}
\begin{picture}(490, 240)
\put(0,0){\includegraphics[width=0.98\linewidth]{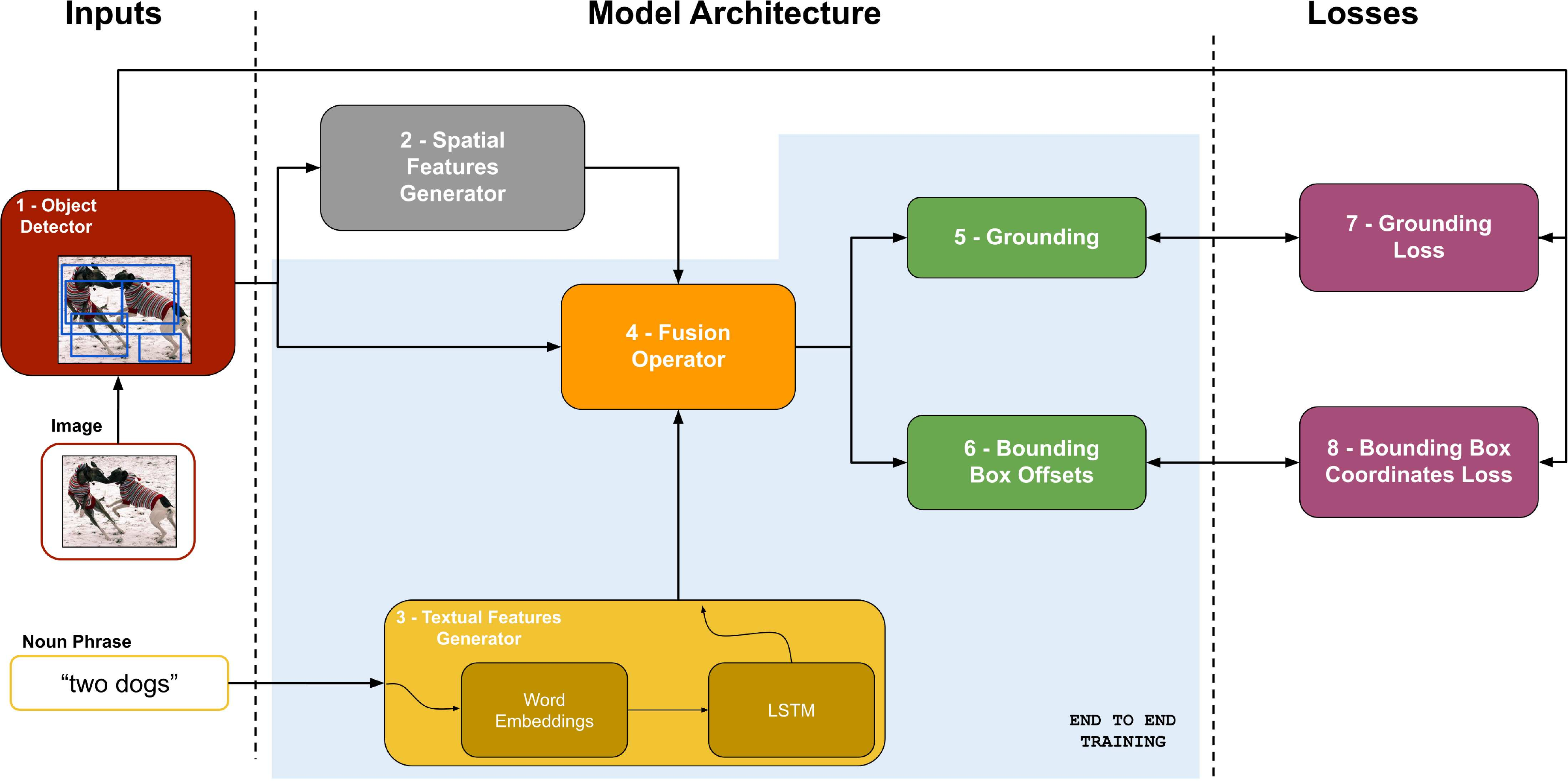}}
% image notation
\put(0, 200){$Pr_{Cls}(\mathcal{P}_{\bI})$}
\put(92, 200){$\mathcal{P}_{\bI}$}
\put(97, 143){$\mathcal{P}_{\bI}, H^v$}
\put(190, 200){$H^s$}
% text notation
\put(95, 35){$W^{\bq}$}
\put(205, 25){$E^{\bq}$}
\put(195, 60){$H^\star$}
% conditional and pred
\put(255, 143){$H^{||}$}
% losses
\put(380, 175){$\mathcal{L}_{g}$}
\put(380, 105){$\mathcal{L}_{c}$}
\end{picture}
\end{center}
\caption{\label{fig:model}Our two-stage model architecture overview. 
(\textbf{1}) Initially, the image is processed by a pre-trained \textit{Faster R-CNN} object detector in order to extract all the proposals bounding boxes from which (\textbf{2}) the spatial features are generated. 
Then, the model (\textbf{3}) generates the textual features from the input noun phrase using the \textit{Textual Features Generator} module, by first retrieving each word embedding and then using an LSTM network. 
Finally, the model (\textbf{4}) fuses together all the visual, spatial, and textual features by the \textit{Fusion Operator}, obtaining new features that are then used in the (\textbf{5}) \textit{Grounding} and (\textbf{6}) \textit{Bounding Box Offsets} modules, respectively.
Our defined losses $\mathcal{L}_{g}$ (\textbf{7}) and $\mathcal{L}_{c}$ (\textbf{8}) are used in order to train the network end-to-end on the components included in the light blue background.}
\end{figure*}

%\begin{center}
%\begin{picture}(490, 240)
%\put(0,0){\includegraphics[width=1\linewidth]{img/model.eps}}
%% image notation
%\put(0, 200){$Pr_{Cls}(\mathcal{P}_{\bI})$}
%\put(92, 192){$\mathcal{P}_{\bI}$}
%\put(97, 137){$\mathcal{P}_{\bI}, H^v$}
%\put(190, 190){$H^s$}
%% text notation
%\put(95, 35){$W^{\bq}$}
%\put(205, 25){$E^{\bq}$}
%\put(195, 60){$H^\star$}
%% conditional and pred
%\put(255, 137){$H^{||}$}
%% losses
%\put(380, 170){$\mathcal{L}_{g}$}
%\put(380, 102){$\mathcal{L}_{c}$}
%\end{picture}
%\end{center}

%\begin{figure*}
%\begin{center}
%\includegraphics[width=\linewidth]{img/model.png}
%\end{center}
%\caption{Our two-stage model architecture overview. 
%(\textbf{1}) Initially, the image is processed with a pre-trained \textit{Faster R-CNN} object detector to %extracts all the proposals bounding boxes from which (\textbf{2}) the spatial features are extracted. 
%Then, the model (\textbf{3}) extracts the textual features from the phrase using the \textit{Textual %Features Extraction Module}, by first retrieving the each word embedding and after using an LSTM network. 
%Finally, the model uses the textual features to (\textbf{4}) condition the bounding boxes proposals with %the \textit{Conditional Module}, which are used both in the (\textbf{5}) \textit{Grounding Module} and the %(\textbf{6}) \textit{Bounding Box Offsets Module}. 
%}
%\label{fig:model}
%\end{figure*}
In this section, we first describe the structure of our model, and then we describe the training procedure, which exploits the
original part of our proposal, e.g. a loss function composed of novel sub-losses.

\subsubsection{Model}
Our model, outlined in Figure~\ref{fig:model}, follows a typical basic architecture for visual-textual grounding tasks. It is based on a two-stage approach in which, initially, a pre-trained object detector is used to extract, from a given image $\bI$, a set of $k$ bounding box proposals $\mathcal{P}_{\bI}$, %=\{\bp_i\}_{i=1}^k$, where $\bp_i \in \mathbb{R}^4$, 
jointly with visual features $H^v$. %=\{\bh^v_i\}_{i=1}^k$,} where $\bh^v_i \in \mathbb{R}^{v}$.
The features represent the internal object detector activation values before the classification layers and regression layer for bounding boxes.
%Moreover, our model extracts the spatial features \mbox{$H^s=\{\bh^s_i\}_{i=1}^k$,} where $\bh^s_i \in \mathbb{R}^{s}$ from all the bounding boxes proposals, where the spatial features for the proposal $\bp_i$ %are defined as:
%\begin{equation}
%    \bh^s_i = \left[ \frac{x1}{wt}, \frac{y1}{ht}, \frac{x2}{wt}, \frac{y2}{ht}, \frac{(x2-x1) \times (y2-y1)}{wt \times ht}\right],
%\end{equation}
%where $(x1, y1)$ refers to the top-left bounding box corner, $(x2, y2)$ refers to the bottom-right bounding box corner, $wt$ and $ht$ are the  width and height of the image, respectively.  
Moreover, our model extracts the spatial features $H^s$ from the proposals. %=\{\bh^s_i\}_{i=1}^k$,} where $\bh^s_i \in \mathbb{R}^{s}$ are the spatial features for the bounding box proposal $\bp_i$. 
% More details regarding the spatial features are reported in the Implementation Details section of the Supplementary Material.
We also assume that the object detector returns, for each bounding box proposal $\bp_i \in \mathcal{P}_{\bI}$, 
a probability distribution $Pr_{Cls}(\bp_i)$ over a set $Cls$ of predefined classes, i.e. the probability for each class $\xi\in Cls$ that the content of the bounding box proposal $\bp_i$ belongs to $\xi$. This information is typically returned by most of the object detectors, and it will be used to define our novel loss terms.

Regarding the textual features extraction, given a noun phrase $\bq_j$, initially all its words $W^{\bq_j}$ % = \{w^{\bq_j}_i\}_{i=1}^l$ 
are embedded in a set of vectors $E^{\bq_j}$. % = \{\be^{\bq_j}_i\}_{i=1}^l$ where $\be^{\bq_j}_i \in \mathbb{R}^{w}$, where $w$ is the size of the embedding.
Then, our model applies a LSTM~\cite{hochreiter1997long} neural network to generate from the sequence of word embeddings only one new embedding  $\bh^\star_j$ for each phrase $\bq_j$.
%This textual features extraction is defined as:
%\begin{equation}
%    \bh^\star_j = L1 \left(LSTM(E^{\bq_j}) \right),
%\end{equation}
%where $\bh^\star_j \in \mathbb{R}^t$ is the LSTM output of the last word in the noun phrase $\bq_j$, and $L1$ is the L1 normalization function.
Once vector  $\bh^\star_j$ has been generated from the noun phrase $\bq_j$, the model performs a multi-modal feature fusion operation in order to combine the information contained in $\bh^\star_j$ with each of the proposal bounding boxes $\bh^v_z\in H^v$.
For this operation, we have decided to use a simple function that merges the multi-modal features together rather than relying on a more complex operator, such as bilinear-pooling or deep neural network architectures. We leave the use of a more complex  fusion operator, that will lead to further improvements, for future work.
The multi-modal fusion component we adopted returns the set of new vectorial representations $H^{||}$.  %=\{\bh^{||}_{jz}\}_{j \in [1,\ldots, m], z\in [1,\ldots, k]}$}, where  vectors $\bh^{||}_{jz}$ are defined as:
%\begin{equation}
%    \bh^{||}_{jz} = LR\left(\bW^{||} \left(\bh^\star_j \,||\, \bh^s_z \,||\, L1(\bh^v_z) \right) + \bb^{||} \right),
%\end{equation}
%where $||$ indicates the concatenation operator, \mbox{$\bh^{||}_{jz} \in \mathbb{R}^{c}$},  $LR$ indicates the leaky-relu activation function, \mbox{$\bW^{||} \in \mathbb{R}^{c\times (t+s+v)}$} is a matrix of weights, and $\bb^{||} \in \mathbb{R}^{c}$ is a bias vector.

Finally, the model predicts the probability $\bP_{jz}$ that a given noun phrase $\bq_j$ is referred to a proposal bounding box $\bp_z$. % as:
%\begin{equation}
%    \bP_{jz} = 
%    \frac{\exp(\bW^g \times \bh^{||}_{jz} + b^g)}
%    {\sum{i=1}{k} \exp{(\bW^g \times \bh^{||}_{ji} + b^g)} },
%\end{equation}
%where $\bW^g \in \mathbb{R}^{1\times c}$ and $b^g \in \mathbb{R}$ are weights. % of a fully connected neural network. 
Indeed, the representations %$\bh^{||}_{jz}$ 
of the proposals bounding box features conditioned with the textual features 
can also be used to 
%the proposal bounding box coordinates extracted by the object detector are not.
%For this reason \textcolor{red}{perche per questa ragione ? non vedo il nesso causale}, we add to the model the possibility to 
refine the proposal bounding box coordinates, that are generated by the object detector independently by
the textual features. Specifically, our model does not predicts new bounding box coordinates, but  offsets for the coordinates. % defined as 
%\begin{equation}
%    \bo_{jz} = \bW^\mathcal{B} \times \bh^{||}_{jz} + \bb^\mathcal{B},
%\end{equation}
%where $\bW^\mathcal{B} \in \mathbb{R}^{4\times c}$ and $\bb^\mathcal{B} \in \mathbb{R}^4$ are a matrix of weights and a bias vector, respectively. % of a fully connected neural network. 
%The final predicted bounding boxes coordinates are then obtained as the sum of the proposal bounding boxes coordinates with the predicted offsets.

Technical details regarding the model are reported in the Supplementary Material.
%%%%%%%%%%%%%%%%%%%%%%%%%%%%%%%%%%%%%%%%%%%%%%%%%%%%%%%%%%%%%%%%%%%%%%%%
%%%%%%%%%%%%%%%%%%%%%%%%%%%%%%%%%%%%%%%%%%%%%%%%%%%%% TRAINING %%%%%%%%%
%%%%%%%%%%%%%%%%%%%%%%%%%%%%%%%%%%%%%%%%%%%%%%%%%%%%%%%%%%%%%%%%%%%%%%%%
\subsubsection{Training}
\label{sec:training}
In this section, we present the main novel contribution of the paper, i.e. a loss function composed of novel terms.
%\textcolor{red}{The general idea is to apply gradient descent to each bounding box in proportion to how much similar its classes probability distribution is with that of the ground truth and how much the two bounding boxes overlap (IoU). Furthermore, in the literature the loss related to the prediction of the bounding box coordinates is calculated only on the proposal with the greatest overlap with the ground truth, while using our semantic information, we perform the gradient descent on a set of bounding boxes.}
The basic idea is to exploit the semantic information associated with bounding box proposals, i.e. the probability distribution over classes of the content of a bounding box returned by the object detector, in both the loss term concerning the grounding and the loss term concerning the refinement of the bounding box coordinates. In fact, differently from most of the previous works that use the \textit{cross-entropy (CE)} loss or the standard \textit{Kullback–Leibler(KL) divergence} loss for grounding, our model implements a KL divergence loss in which the ground truth probability is built also considering $Pr_{Cls}(\bp_i)$ with $\bp_i\in \mathcal{P}_{\bI}$. Moreover, regarding the bounding boxes coordinates refinement, differently from previous works that use the \textit{Smooth\textsubscript{L1} loss}, our model adopts the \textit{CIoU loss}~\cite{zheng2020enhancing}.
To the best of our knowledge, this is the first work adopting the \textit{CIoU} loss in order to refine the final bounding boxes coordinates. Another difference with respect to  all the refinement losses proposed in the literature is that we do not restrict the coordinates refinement only to the best proposal coordinates, but we extend the refinement to the subset of proposals that  significantly overlap (according to an hyper-parameter) the ground truth, modulating the refinement by the agreement between the class probability of the best proposal and the class probability of the considered proposal. 
For the sake of presentation, we formally define the new loss terms in the following referring to a single example. The total loss is then obtained by summing up the contributions of all examples in the training set.

Given a training example $(\bI, \mbox{S}, \Gamma^{gt})$, and the bounding box proposals set $\mathcal{P}_{\bI}$, we define the loss function $\mathcal{L}$ (for a single example) as:
\begin{equation*}
    \mathcal{L} = \mathcal{L}_{g}(\bP, \mathcal{P}_{\bI}, \Gamma^{gt}) + \lambda \mathcal{L}_{c}(\mathcal{P}_{\bI}, \Gamma^{gt}),
\end{equation*}
where $\mathcal{L}_{g}$ is the loss used to ``shape'' the grounding  distribution of %bounding box 
proposals for each specific query in input, i.e. the probability that a given proposal is associated to a given query, $\mathcal{L}_{c}$ is the loss related to the refinement of the bounding boxes coordinates, and $\lambda$ is a trade-off parameter.

%Differently from most of the previous works that use the \textit{cross-entropy (CE)} loss or the standard \textit{Kullback–Leibler(KL) divergence} loss for grounding, our model implements a KL divergence loss in which the ground truth probability is built also considering $Pr_{Cls}(\bp_i)$ with $\bp_i\in \mathcal{P}_{\bI}$, i.e. the semantic information associated to bounding box proposals.
% Formally, given a ground truth bounding box $\bb^{gt}$ for a given phrase $\bq \in Q$ and the bounding box proposals $\mathcal{P}_{\bI}$, we define  the vector $\bu$ as:
Specifically, given $m$  the number of noun phrases and $k$ the number of bounding box proposals, we define the entries $(j \in [1,\ldots, m]$, $z \in [1,\ldots, k])$ of matrix $\bU$ as
%\begin{align}
    $\bU_{jz} = IoU(\bb_j^{gt}, \bp_{z})$ where $(\bq_j^{gt}, \bb_j^{gt}) \in \Gamma^{gt}$,
%\end{align}
the best proposal bounding box as $\bp_{j*}$ where
%\begin{align}
   $ j* = argmax_{z \in [1, ..., k]} \, \bU_{jz}$, 
%\end{align}
and the entries $(j \in [1,\ldots, m]$, $z \in [1,\ldots, k])$ of matrix $\bC$ containing the cosine similarity scores among the predicted class probabilities of the bounding box proposals as
%\begin{align}
    \mbox{$\bC_{jz} = Sim\left(Pr_{Cls}(\bp_{j*}), Pr_{Cls}(\bp_{z}) \right)$,}
%\end{align}
where $Sim$ is the cosine similarity function. Given these definitions, we can define
the entries of the target probability 
$\bP^{target}$ as:
\begin{align*}
    \bP^{target}_{jz} &= \frac{\bU^*_{jz} }{\sum{i=1}{k}\bU^*_{ji} }, \mbox{ where}
\end{align*}
\begin{align*}
\bU^*_{jz} &=  
    \begin{cases} 
    \bU_{jz} \bC_{jz},    & if \  \   \bU_{jz} \geq \eta \\
    0,                      &    otherwise
    \end{cases},
\end{align*}

\noindent and $\eta$ is a predefined threshold, i.e. an hyper-parameter.

On the basis of the above definitions, we define the grounding loss as:
\begin{align*}
    \mathcal{L}_{g}(\bP, \mathcal{P}_{\bI}, \Gamma^{gt})
    &= \frac{1}{m} \sum{j=1}{m} KL_{div}(\bP_{j} || \bP^{target}_{j}),\\  
    &= \frac{1}{m}\sum{j=1}{m} \sum{z=1}{k} \bP_{jz} \log \left( \frac{\bP_{jz}}{\bP^{target}_{jz}} \right),
\end{align*}
where $KL_{div}$ is the KL divergence function, $\bP_{j}$ ($\bP^{target}_{j}$) is the $j$-th row of $\bP$ ($\bP^{target}$),  %$m$ is the number of noun phrases, 
%$k$ is the number of proposal bounding boxes, 
and $\bP_{jz}$ is the model predicted probability that the noun phrase \mbox{$\bq_j \in Q$} refers to the image content localized by $\bp_z \in \mathcal{P}_{\bI}$.  

Indeed, the grounding loss captures both the bounding box spatial information and the semantic information determined by the bounding box classes. 
Whenever a bounding box is located near the ground truth bounding box and its class probability distribution is similar to the one of the best proposal $\bp_{j*}$, then the loss favours the prediction of the bounding box, otherwise the loss penalizes the bounding boxes according to their different probability distribution and spatial location.
Previous works exploiting the KL divergence aims to maximize the probability of a proposal bounding box just considering their spatial location.
% This loss increases the accuracy of our model by 0.2\% if compared to the standard cross-entropy loss, where  $\bP^{target}$ is the one hot vector in which the proposal $\bp_*$ has probability equal to 1:

We now define the novel refinement loss. In order to do that, given a query $\bq_j$, we need to define the following subset  $\mathcal{S}_j \subseteq \mathcal{P}_{\bI}$ of proposals:
%XXXXX
%Regarding the bounding boxes coordinates refinement, differently from previous works that use the \textit{Smooth\textsubscript{L1} loss}, our model adopts the \textit{CIoU loss}~\cite{zheng2020enhancing}.
%To the best of our knowledge, this is the first work adopting the \textit{CIoU} loss in order to refine the final bounding boxes coordinates.
%Moreover, differently from all the works in the literature in which the coordinates refinement loss is computed only on the best proposal $\bp_{j*}$ coordinates, our loss, given a query $\bq_j$, penalizes the subset $\mathcal{S}_j \subseteq \mathcal{P}_{\bI}$ defined as:
\begin{align*}
    \mathcal{S}_j = \{\bp_z \mid \bp_z \in \mathcal{P}_{\bI} \land \bU^*_{jz} \ge 0 \},
\end{align*}
%Specifically, given a query $\bq_j \in Q$, 
which allows us to define our loss $\mathcal{L}_{c}$ as:
\begin{align*}
    \mathcal{L}_{c}(\mathcal{P}_{\bI}, \Gamma^{gt}) &=
    \frac{1}{m}\sum{j=1}{m}\sum{\bp_z \in \mathcal{S}_j}{} \hat{\bU}_{jz}\mathcal{L}_{CIoU}(\bp_z, \bb^{gt}_j),
\end{align*}
where $(\bq_j^{gt}, \bb_j^{gt}) \in \Gamma^{gt}$, and 
\begin{align*}
    \hat{\bU}_{jz}  &= \frac{\bU^*_{jz}}{max_{z \in [1, k]}\bU^*_{jz} + \epsilon},
\end{align*}
in which $\epsilon$ is a small value added to avoid division by 0, and $max_{z \in [1, k]}$ is the maximum function applied along the indexes $z \in [1, k]$.
Intuitively, for each bounding box proposal which overlaps with the ground truth (according to the parameter $\eta$), this loss refines the coordinates proportionally to the ``semantic'' of the bounding box.
Note that adopting the normalized scores $\hat{\bU}_{jz}$, the model does not penalize the loss on the best proposal bounding box $j*$.

We would like to highlight that our work is the first proposing the exploitation of the probabilities distributions over the object detector classes to address the supervised visual grounding task.
However, in weakly-supervised visual-textual grounding ({\it not our task}) some works (e.g. \cite{DBLP:conf/iccv/WangS19a}) leverage the information of the bounding box class with the {\it highest} probability.

\section{Experimental Assessment}
\label{sec:experiments}
We have compared our model results on two widely adopted datasets (i.e., Flickr30k Entities and ReferIt) considering several competing approaches in the literature, including state-of-the-art models.
In addition to that, in order to prove the usefulness of our losses independently by our model architecture, we have also adopted our losses on the
 DDPN model. The choice of this model was due to: (i) publicly available code\footnote{We have adapted the official code: \url{https://github.com/XiangChenchao/DDPN}.};  (ii) published results on both Flickr30k Entities and ReferIt datasets, with state-of-the-art results on ReferIt; (iii) and exploitation of the same object detector used in our work. 
 
\subsection{Datasets and Evaluation Metric}
%\textcolor{red}{Sezione modificata}
Flickr30k Entities and ReferIt constitute the two most common datasets used in the literature, although other datasets have been used (e.g., \cite{Chen_2020_CVPR, kazemzadeh2014ReferItgame, yu2016modeling, mao2016generation}).
% \paragraph{Flickr30K}
% The Flickr30K Entities dataset~\cite{flickrentitiesijcv, flickr30k}  contains $32$K images, $275$K bounding boxes, $159$K sentences, and $360$K noun phrases. 
% Each image is associated with five sentences with a variable number of noun phrases, and each noun phrase is associated with a set of bounding boxes ground truth coordinates.
% Following all works in the literature, if a noun phrase corresponds to multiple ground truth bounding boxes, we merged the boxes and used their union region as its ground-truth.
% On the contrary, a noun phrase with no associated bounding box was removed from the dataset.
% We used the standard split for training, validation, and test set as defined in~\cite{flickrentitiesijcv}, consisting  of $30$K, $1$K, and $1$K images, respectively.
% \paragraph{ReferIt}
% The ReferIt~\cite{KazemzadehOrdonezMattenBergEMNLP14} dataset contains $20$K images, $99$K bounding boxes, and $130$K noun phrases. 
% This dataset differs from Flickr30k since it does not contain sentences, which means that the noun phrases are mutually independent.
% feature fusion operator that assumes the presence of the input sentence containing all the noun phrases, cannot be applied to it.
% We used the same split as in~\cite{flickrentitiesijcv} that consists of $9$K images of training, $1$K images of validation, and $10$K images of test.\\
The Flickr30k Entities dataset~\cite{plummer2015flickr30k, flickr30k}  contains $32$K images, $275$K bounding boxes, $159$K sentences, and $360$K noun phrases. 
The ReferIt~\cite{KazemzadehOrdonezMattenBergEMNLP14} dataset contains $20$K images, $99$K bounding boxes, and $130$K noun phrases. 
This dataset differs from Flickr30k Entities since it does not contain sentences, which means that the noun phrases are mutually independent.
%For this reason, the state-of-the-art models that depend on a sentence linking all the noun phrases, since they use a 
%feature fusion operator that assumes the presence of the input sentence containing all the noun phrases, cannot be applied to it.
We refer the reader to the Datasets Details section of the Supplementary Material for more details.

Aligned with the works in the literature, we adopted the standard \textit{Accuracy} metric.
%and the \textit{point game accuracy} that is used in recent works~\textcolor{red}{aggiungi riferimenti}.
Given a noun phrase, it considers a bounding box prediction to be correct if and only if the intersection over union value between the predicted bounding box and the ground truth bounding box is at least $0.5$.
%while the latter accuracy metric considers the prediction right if and only if the center of the predicted bounding box is contained in the ground truth bounding box.
%Indeed, the standard accuracy metric considers both the geometric size of the bounding boxes and their location in the image, while the point game accuracy focus only in the bounding boxes location.
%\begin{figure*}[t]
%\begin{center}
%\includegraphics[width=0.65\linewidth]{img/Figure_small.png} % era 0.93
%\end{center}
%\caption{\label{fig:quality}A qualitative example of our model on the test image id: %$23016347$.
%The bounding boxes associated with each query are reported in the left picture, %while the ground truth is reported in the right picture. The complete sentence in %input is reported at the bottom of the figure. The prediction for the query ``a %tennis ball'' is evaluated as wrong, even if the bounding box is very close to the %ground truth. }
%\end{figure*}

\subsection{Model Selection and Implementation Details}
To evaluate our model on the test set of Flickr30k Entities and ReferIt datasets, we have chosen the epoch in which the model achieved the best \textit{Accuracy} metric on the validation set.
We have performed a grid search for the best hyper-parameters mainly for the Flickr30k Entities dataset, ad exception of the losses hyper-parameters visible in Section \ref{sec:ablation}. For the ReferIt dataset, we have used the other hyper-parameters values selected on the Flickr30k Entities dataset.
We have used the Adam optimizer with exponential learning rate scheduler set to $0.9$, and the following values for the learning rate: $\{0.05, 0.03, 0.01, 0.005, 0.001\}$, $c: \{2048, 2053, 2060\}$, and $\eta: \{0.1, 0.3, 0.4, 0.45, 0.5, 0.55\}$. % and, we have tried to learn the pre-trained GloVe embeddings.
Other hyper-parameters are fixed to single values. For the textual features: \mbox{$w=300$}, $t=500$, and the LSTM network uses only one hidden layer of dimension $t$.
For the image features, we have extracted a fixed number $k=100$ of proposals for each image, \mbox{$v=2048$} from the ResNet-101's layer \emph{pool5\_flat}, and $s=5$.
In both datasets, we have found that the best model \textit{Accuracy} is achieved at epoch $9$ of training with learning rate set to $0.001$ and $c=2053$.
For Flickr30k Entities we have set $\eta = 0.3$ and $\lambda = 1$, while for ReferIt we have set $\eta = 0.5$ and $\lambda = 1.4$.
The code is publicly available on GitHub~\footnote{\url{https://github.com/drigoni/Loss_VT_Grounding}}.
We refer the reader to the Implementation Details section of the Supplementary Material for more details.

\subsection{Results}
\begin{table}
\small
\centering
\caption{Results obtained on Flickr30k test set. 
\textit{Accuracy} indicates in percentage the standard accuracy metric.
All values are copied from the original articles.
"*" indicates that the reported model accuracy is referring to the version of the model in their ablation study, since the complete model uses query dependency information that we do not exploit.
}
\begin{tabular}{| c | c |}
\hline
\textbf{Model} & \textbf{Accuracy (\%)} \\
\thickhline
SCRC~\cite{hu2016natural}                       & 27.80  \\
SMPL~\cite{wang2016structured}                  & 42.08  \\
NonlinearSP~\cite{wang2016learning}             & 43.89  \\
GroundeR~\cite{rohrbach2016grounding}           & 47.81  \\
MCB~\cite{DBLP:conf/emnlp/FukuiPYRDR16}         & 48.69  \\
RtP~\cite{plummer2015flickr30k}                 & 50.89  \\
Similarity Network~\cite{wang2018learning}      & 51.05  \\
IGOP~\cite{DBLP:conf/nips/YehXHDS17}            & 53.97  \\
SPC+PPC~\cite{DBLP:conf/iccv/PlummerMCHL17}     & 55.49  \\
SS+QRN~\cite{chen2017query}                     & 55.99  \\
SeqGROUND~\cite{dogan2019neural}                & 61.60  \\
CITE~\cite{plummer2018conditional}              & 61.89  \\
QRC net~\cite{chen2017query}                    & 65.14  \\
YOLO~\cite{yang2019fast}                        & 68.69   \\
%~\cite{plummer2020revisiting}\textcolor{red}{bho}                   & 71.90  \\
%~\cite{gupta2020contrastive}\textcolor{red}{bho}                    & 51.67  \\
DDPN~\cite{DBLP:conf/ijcai/YuYXZ0T18}           & 73.30   \\
CMGN~\cite{DBLP:conf/aaai/LiuWZH20}*            & 73.46   \\
SL-CCRF~\cite{DBLP:conf/emnlp/LiuH19}           & 74.69   \\
%CMGN-Full~\cite{DBLP:conf/aaai/LiuWZH20}        & 76.74   \\

\thickhline
Ours                                            & \textbf{75.55} \\
DDPN~\cite{DBLP:conf/ijcai/YuYXZ0T18}    using our losses     & 74.33 \\
\hline
\end{tabular}
\label{tab:results_flickr30k}
\end{table}

Table~\ref{tab:results_flickr30k} reports the results obtained on the Flickr30k Entities dataset by our approach and many other approaches presented in the literature, including the most recent state-of-the-art models reported at the bottom part of the table. 
Concerning the model CMGN developed in~\cite{DBLP:conf/aaai/LiuWZH20}, for the sake of a fair comparison, we have reported the performance obtained using the same setting of our model. In fact, the complete version of the CMGN model achieves an \textit{Accuracy} of $76.74\%$, but exploiting query dependency information that we could exploit as well. The integration of this information in our model is left for future work. It can be noted that our approach significantly improves over competing approaches. Moreover, the DDPN model where our losses are used (last row of the table) shows a significant improvement  in performance ($1.03\%$) with respect to the original version.

\begin{table}
\small
\centering
\caption{Results obtained on ReferIt test set. 
\textit{Accuracy} indicates in percentage the standard accuracy metric.
All values are reported from the original articles.
}
\begin{tabular}{| c | c |}
\hline
\textbf{Model} & \textbf{Accuracy (\%)} \\
\thickhline
SCRC~\cite{hu2016natural}                       & 17.93   \\
GroundeR~\cite{rohrbach2016grounding}           & 26.93   \\
MCB~\cite{DBLP:conf/emnlp/FukuiPYRDR16}         & 28.91   \\
CITE~\cite{plummer2018conditional}              & 34.13   \\
IGOP~\cite{DBLP:conf/nips/YehXHDS17}            & 34.70   \\
\cite{wu2017end}                                & 36.18   \\
QRC net~\cite{chen2017query}                    & 44.10   \\
\cite{DBLP:conf/mm/LiWLZLXF17}                  & 44.20   \\
%~\cite{plummer2020revisiting}\textcolor{red}{bho}                   & 57.80    \\
\cite{yang2019fast}                             & 59.30   \\
DDPN~\cite{DBLP:conf/ijcai/YuYXZ0T18}           & 63.00   \\
\thickhline
Ours                                            & 66.02 \\
DDPN~\cite{DBLP:conf/ijcai/YuYXZ0T18}  using our losses  & \textbf{66.66} \\
\hline
\end{tabular}
\label{tab:results_referit}
\end{table}

Table~\ref{tab:results_referit} reports the results obtained on the ReferIt dataset by our approach and the subset of the  competing approaches reported in Table~\ref{tab:results_flickr30k} that can be applied to this dataset, plus additional approaches that have been assessed on this dataset\footnote{Some of them do not define an acronym, so we just use the reference to the paper. }. 
Our model improves the \textit{Accuracy} value by $3.02$\% when compared to the state-of-the-art model (i.e., DDPN) for this dataset, representing a more significant gain than the one obtained on Flickr30k Entities. On the other hand, adopting our losses in DDPN leads to the best performance, with an improvement over the original version of $3.66\%$.
In the ReferIt dataset, each sentence corresponds to a single query independently from the others.
In contrast, in Flickr30k Entities, a sentence could contain more queries that are semantically related among them. 
For this reason, models that apply complex multi-modal feature fusion components that aim to capture information among the queries extracted by the sentence in input sometimes do not consider the ReferIt dataset.
Thus, the set of the models used as comparison in the ReferIt dataset is not the same as in Flickr30k Entities and these reasons could explain the higher gain in \textit{Accuracy} obtained in ReferIt than Flickr30k Entities.

We have also calculated the \textit{Point Game Accuracy} which is recently used for a few models addressing the weakly-supervised task. 
It considers a prediction to be correct if and only if the center of the predicted bounding box is contained in the ground truth bounding box.
In particular, our model obtains $87.96$\%  and $78.0$\% on Flickr30k Entities and ReferIt, respectively.
These values are far better than the ones reported in the literature, and they suggest that a significant subset of predictions that are considered to be wrong according to the \textit{Accuracy} metric, still refer to bounding boxes that have a significant overlap with the ground truth.
%Future studies could deepen this two cases in order to address explicitly these two scenarios.

%Finally, to prove the potential of our losses independently by our model architecture, we have implemented our losses on a state-of-the-art model.
%In particular, we have exploited our losses on the DDPN model\footnote{We have reproduced the model following the official code: \url{https://github.com/XiangChenchao/DDPN}.} as: (i) it is applied on both Flickr30k and ReferIt datasets; (ii) it presents state-of-the-art results on ReferIt; (iii) and its object detector is the same used in our work. We achieved an \textit{Accuracy} of $74.33$\% and $66.66$\% on the Flickr30k and ReferIt datasets, respectively, that are higher results than those reported by the authors.

\subsubsection{Qualitative Results}
\begin{figure}
\begin{center}
\includegraphics[width=1\linewidth]{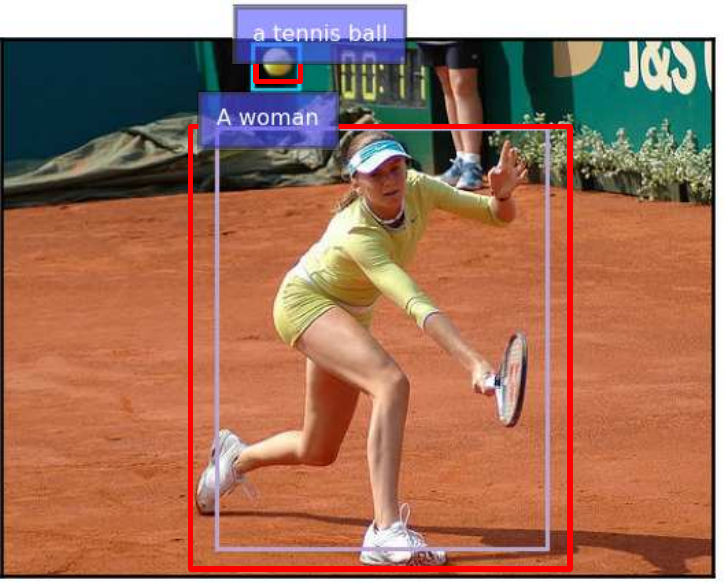}
\includegraphics[width=1\linewidth]{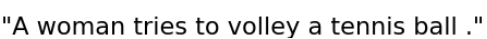}
\end{center}
\caption{\label{fig:quality1}This picture reports a qualitative example of our model on the Flickr30k test image id: $23016347$.
%The predicted bounding boxes associated with each query are reported in \textbf{light blue}, while the ground truth is reported \textbf{red}. The complete sentence in input is reported at the bottom of the figure. 
The ground truth bounding boxes associated with each query are reported in \textbf{red}.The prediction for the query ``a tennis ball'' is evaluated as wrong, even if the bounding box is very close to the ground truth.}
\end{figure}

\begin{figure}
\begin{center}
\includegraphics[width=1\linewidth]{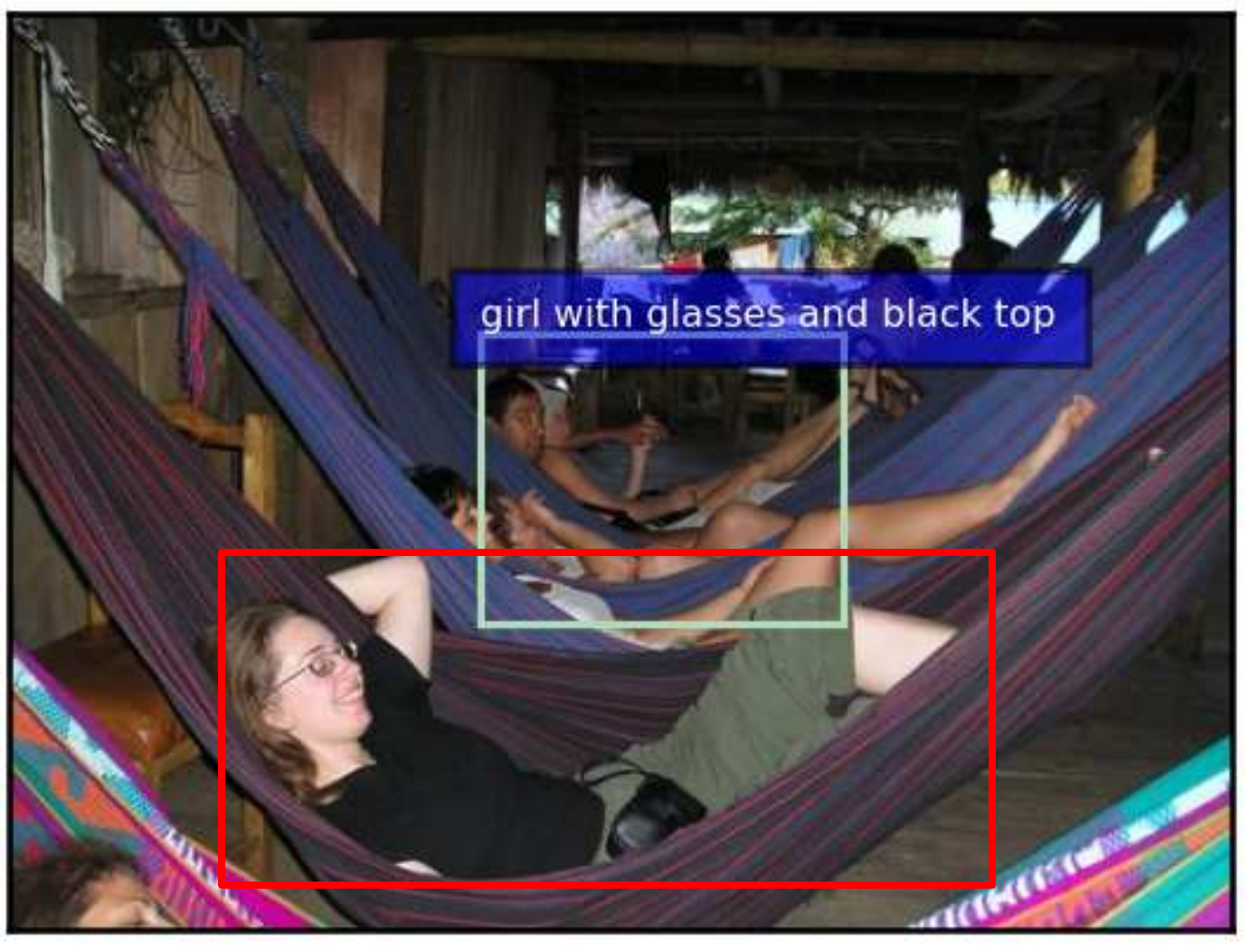}
\includegraphics[width=0.84\linewidth]{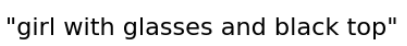}
\end{center}
\caption{\label{fig:quality2}This picture reports a qualitative example of our model on the ReferIt test image id: $14651$ .
The ground truth bounding box is reported in \textbf{red}. The complete sentence in input is reported at the bottom of the figure. The predicted bounding box presents an intersection over union value with the ground truth of 0.08.}
\end{figure}

\begin{figure}
\begin{center}
\includegraphics[width=0.75\linewidth]{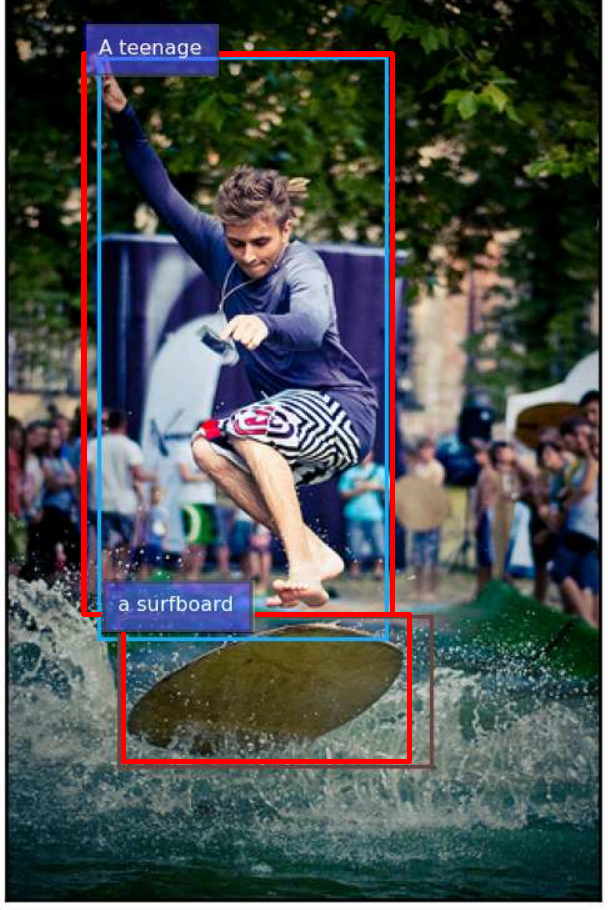}
\includegraphics[width=0.8\linewidth]{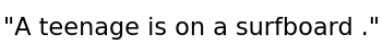}
\end{center}
\caption{\label{fig:quality3}This picture reports a qualitative example of our model on the Flickr30k test image id: $6059154572$.
The ground truth bounding boxes associated with each query are reported in \textbf{red}. The complete sentence in input is reported at the bottom of the figure. All bounding boxes are predicted correctly. }
\end{figure}

Figures~\ref{fig:quality1},\ref{fig:quality2}, and \ref{fig:quality3} show qualitative examples predicted by our model on the test set of both Flickr30k Entities and ReferIt datasets. 
%The example considers a Flickr30K Entities test image with one of its five sentences.
%For each query extracted from the selected sentence reported at the bottom of the figure, on the left picture its predicted bounding box is reported. Similarly,  on the right picture its ground truth is reported as well. 
In our model predictions, we have noticed that when the query refers to a small object in the image, most of the time our model predicts a very close bounding box, but not enough to have the IoU score over the $0.5$ value.
This is the case for the query ``a tennis ball'' in the figure~\ref{fig:quality1}.
More examples are reported on the Qualitative Results section of the Supplementary Material.
%\begin{figure}
%\begin{center}
%\includegraphics[width=1\linewidth]{img/qualitative_examples/q3_overlap.pdf}
%\includegraphics[width=0.65\linewidth]{img/qualitative_examples/q3_query.pdf}
%\end{center}
%\caption{\label{fig:quality}This picture reports a qualitative example of our model on the ReferIt test image id: $8071$ .}
%\end{figure}
%\begin{figure}
%\begin{center}
%\includegraphics[width=1\linewidth]{img/qualitative_examples/q5_overlap.pdf}
%\includegraphics[width=0.5\linewidth]{img/qualitative_examples/q5_query.pdf}
%\end{center}
%\caption{\label{fig:quality}This picture reports a qualitative example of our model on the ReferIt test image id: $22698$.}
%\end{figure}

%Img: 23016347 (500x377)
%Sentence:  A woman tries to volley a tennis ball .
%GT bounding boxes:  [[137.  64. 371. 366.]
% [182.   6. 208.  27.]]
%Phrase:  A woman
%Pred:  [151.  63. 383. 358.]
%Prob:  0.4779538
%iou:  0.8690202
%Phrase:  a tennis ball
%Pred:  [176.   3. 209.  36.]
%Prob:  0.99647313
%iou:  0.4989908
%Img: 707941195 (334x500)

\subsubsection{Ablation Study}
\label{sec:ablation}
\begin{table*}
\small
\centering
\caption{Accuracy obtained on Flicker30k Entities and ReferIt datasets as the losses functions and hyper-parameters values change.
\textit{CE} indicates the cross-entropy loss, \textit{SmoothL1} indicates the Smooth L1 loss, \textit{KL-Sem} indicates our KL loss with the semantic information and \textit{CIoU-Sem} indicate our Complete IoU loss with the semantic information. The baseline model does not use the $\eta$ parameter.}
\centering
\begin{tabular}{| c | c | c | c | c | c | c | c |}
\hline
\multicolumn{2}{|c|}{\textbf{Losses}}&\multicolumn{2}{|c|}{\textbf{Hyper-par.}}&\multicolumn{2}{|c|}{\textbf{Flickr30k (\%)}}  &\multicolumn{2}{|c|}{\textbf{ReferIt (\%)}}   \\
\hline
\textbf{Gr.} &\textbf{Reg.} &\bm{$\lambda$}&\bm{$\eta$}&\textbf{Val.} &\textbf{Test} &\textbf{Val.} &\textbf{Test} \\
\thickhline
\multirow{3}{*}{CE} &\multirow{3}{*}{SmoothL1} 
&0.8     &/       &\textbf{71.25}    &\textbf{71.82}            &64.24              &\textbf{61.81}      \\
&&1      &/      &71.08             &71.61                      &64.19              &61.29               \\
&&1.2    &/      &71.18             &71.21                      &\textbf{64.65}     &61.64   \\
\hline
\multirow{4}{*}{KL} &\multirow{4}{*}{SmoothL1} 
&0.8     &0.4       &71.51          &72.06                  &63.58          &61.38                      \\
&&0.8    &0.5       &72.16          &\textbf{72.55}         &64.57          &\textbf{62.69}             \\
&&1      &0.4       &71.76          &72.34                  &63.93          &61.65                      \\
&&1      &0.5       &\textbf{72.58} &72.18                  &\textbf{64.82} &62.49        \\
\hline
\multirow{4}{*}{KL-Sem} &\multirow{4}{*}{SmoothL1} 
&0.8    &0.4    &72.22          &72.72          &64.38          &61.78                          \\
&&0.8   &0.5    &72.42          &72.41          &64.99          &62.12                          \\
&&1     &0.4    &\textbf{72.54} &\textbf{72.88} &65.04          &62.47              \\
&&1     &0.5    &72.34          &72.83          &\textbf{65.45} &\textbf{62.72}     \\
\hline
\multirow{4}{*}{CE} &\multirow{4}{*}{CIoU-Sem} 
&0.8    &0.4    &73.99          &74.56          &\textbf{67.66} &\textbf{65.47}     \\
&&0.8   &0.5    &73.60          &74.24          &67.41          &65.07                          \\
&&1     &0.4    &\textbf{74.07} &\textbf{74.82} &67.60          &65.42              \\
&&1     &0.5    &73.90          &74.24          &67.24          &65.15                          \\
\hline
\multirow{8}{*}{KL-Sem} &\multirow{8}{*}{CIoU-Sem}     
&0.6    &0.5    &75.17          &75.38          &68.23              &66.31                  \\
&&0.8   &0.5    &75.27          &\textbf{75.67} &68.70              &66.12                  \\
&&1     &0.5    &75.41          &75.53          &68.72              &66.52                  \\
&&1.2   &0.5    &75.23          &75.34          &68.88              &66.37                  \\
&&1.4   &0.5    &75.13          &75.36          &\textbf{68.97}     &66.02      \\
&&1     &0.3    &\textbf{75.60} &75.55          &68.64              &66.49                  \\
&&1     &0.4    &75.40          &75.64          &68.56              &\textbf{66.54}         \\
&&1     &0.6    &74.48          &74.68          &68.02              &65.31                  \\
\hline
\end{tabular}
\label{tab:ablation}
\end{table*}

% # referit epoch 7
% valid: 62.8928
% test: 60.90

% # flickr30k epoch 7
% valid: 70.1240
% test: 71.3141
Our loss is composed by two main components and by two hyper-parameters. Here, we report the contribution of each part of the loss using different hyper-parameters values.
We have performed a set of experiments where the grounding component is alternatively the cross-entropy, the KL divergence or the proposed semantic KL divergence, and the regression component is alternatively the Smooth L1 or the proposed semantic CIoU. 
Moreover, different values for the hyper-parameters are considered. 
The obtained results (Table \ref{tab:ablation}) show that the major contribution to the improvement is given by the \textit{Complete IoU loss with semantic information}, which improves the model \textit{Accuracy} by $\sim 2.6\%$ and $\sim 3.9\%$ on Flickr30K Entities and ReferIt datasets, respectively. 
Significant improvements are also obtained by using the semantic KL divergence in place of cross-entropy or the CIoU-Sem instead of the standard CIoU.
%We can see that using our \textit{KL loss with semantic information} instead of the \textit{cross-entropy loss}, the model \textit{Accuracy} improves by $\sim 1.2\%$ and $\sim 1.4\%$ on Flickr30k and ReferIt datasets, respectively.
%Moreover, our \textit{Complete IoU loss with semantic information} instead of the \textit{SmoothL1 loss}, improves the model \textit{Accuracy} by $\sim 2.6\%$ and $\sim 3.9\%$ on Flickr30k and ReferIt datasets, respectively.
%Using our two losses together, the model \textit{Accuracy} still increase to State-of-the-Art results.
Moreover, results show that our approach is not much sensitive with respect to the hyper-parameters values, and, more importantly, the \textit{Accuracy} on the validation set indeed represents well the \textit{Accuracy} on the test set on both datasets.

%\textbf{\textcolor{brown}{R3}:}
%Very often, the code for the models in the literature are not completely reproducible, because they either not include the pre-processing code or use old software. 
%For these reasons, we have not implemented our losses on previously models.
%Regarding the number of free parameters, we would like to highlight that our model has {\it more free parameters} than many other models presented in literature. In fact, our model architecture is basically the one presented in [50] (in paper), where the latent dimension of the fusion layer is $2053$ instead of $500$. 
%This implies that both the classification (grounding) and regression blocks have an increased input dimension (from $500$ to $2053$).\\
%Our model uses the same dimension for the visual features and the word embedding features in input, the same Adam optimizer with the same hyper-parameters and learning rate as in [50](paper references). 
% la cosa che cambia e' una funzione di attivazione, due l1 al posto delle l2 e lo stato finale dell'output della LSTM della query, in cui noi usiamo 500 e loro 1024 che venogno concatenate alle feature visuali prima della fusione.
%Use of  semantic information: our work is the first proposing the  exploitation of the probabilities distributions over the object detector classes.
%, especially in the regression loss. 
%Usually, the object detector is used just to extract the visual features of the bounding boxes, which are then used to solve the visual-textual grounding. 

\section{Conclusion and Feature Work}
\label{sec:conclusion}
This paper introduced a novel loss for Visual-Textual Grounding, jointly with a simple two-stage approach model.
The novel loss combines a grounding loss and a bounding box coordinates refinement loss, both based on semantic information,
i.e. a probability distribution over a set of pre-defined classes,
returned by the object detector. The experimental assessment showed that the proposed approach was able to reach
a higher accuracy than state-of-the-art models, even without using a more complex multi-modal feature fusion component. 
Specifically, we have compared our results versus several models in the literature over two commonly used datasets, Flickr30K Entities and ReferIt. 
With respect to the best state-of-the-art approaches, on the Flickr30K Entities dataset, we obtained an improvement of $0.86$\%, while on the ReferIt dataset, our model improved the state-of-the-art performance by $3.02$\%. By applying the proposed loss to the DDPN model we were able to significantly improve the performance of the model on both datasets, demonstrating its usefulness independently from the proposed model.

Since this model uses a simple multi-modal feature fusion component, there is space for trivial improvements, including a more sophisticated multi-modal feature fusion component, such as bilinear-pooling and deeper architectures, as well as the exploitation of dependencies among the queries contained by the input sentence.
Future work will also address more sophisticated object detectors, and the idea to include different forms of information, such as a scene graph and prior knowledge.

%\section{Supplementary Material}
%\label{sec:suppl}
%We have reported all the supplementary material available at the following address: \url{https://www.math.unipd.it/~drigoni/files/SAC_2022_A_Better_Loss_for_Visual_Textual_Grounding_Supplementary.pdf}.

%%%%%%%%% REFERENCES
\bibliographystyle{ACM-Reference-Format}
\bibliography{bibliography} 

\end{document}

\title{A Better Loss for Visual-Textual Grounding}
%\titlenote{Produces the permission block, and copyright information}
\subtitle{Supplementary Material}
%\subtitlenote{The full version of the author's guide is available as \texttt{acmart.pdf} document}
  
\renewcommand{\shorttitle}{A Better Loss for Visual-Textual Grounding}

\author{Davide Rigoni}
% \authornote{Dr.~Trovato insisted his name be first.}
\orcid{1234-5678-9012}
\affiliation{%
  \institution{University of Padua \\ Bruno Kessler Foundation}
  \streetaddress{Via Trieste, 63}
  \city{Padua} 
  \state{Italy} 
  \postcode{35121}
}
\email{drigoni@fbk.eu}

\author{Luciano Serafini}
% \authornote{The secretary disavows any knowledge of this author's actions.}
\affiliation{%
  \institution{Bruno Kessler Foundation}
  \streetaddress{Via Sommarive, 18}
  \city{Povo} 
  \state{Italy} 
  \postcode{38123}
}
\email{serafini@fbk.eu}

\author{Alessandro Sperduti}
%\authornote{This author is the one who did all the really hard work.}
\affiliation{%
  \institution{University of Padua}
  \streetaddress{Via Trieste, 63}
  \city{Padua} 
  \state{Italy}
  \postcode{35121}}
\email{sperduti@unipd.it}

%\renewcommand{\shortauthors}{D. Rigoni et al.}

%%%%%%%%% ABSTRAC
\begin{abstract}
In this paper, we provide supplementary material for the paper “A Better Loss for Visual-Textual Grounding”, which has been accepted to be presented at the 37th ACM/SIGAPP Symposium On Applied Computing.
\end{abstract}

\begin{CCSXML}
<ccs2012>
   <concept>
       <concept_id>10010147.10010178.10010224.10010245.10010251</concept_id>
       <concept_desc>Computing methodologies~Object recognition</concept_desc>
       <concept_significance>500</concept_significance>
       </concept>
   <concept>
       <concept_id>10010147.10010178.10010224.10010245.10010250</concept_id>
       <concept_desc>Computing methodologies~Object detection</concept_desc>
       <concept_significance>500</concept_significance>
       </concept>
 </ccs2012>
\end{CCSXML}

\ccsdesc[500]{Computing methodologies~Object recognition}
\ccsdesc[500]{Computing methodologies~Object detection}

%%%%%%%%% KEYWORDS
\keywords{Computer Vision, Visual Textual Grounding, Semantic Loss}

\maketitle

%%%%%%%%% CHAPTERS
\section{Background}
%In order to explain our work, w
We use the following notation:
lower case symbols for scalars and indexes, e.g. $n$;
italics upper case symbols for sets, e.g. $A$;
upper case symbols for textual sentences, e.g. S;
bold lower case symbols for vectors, e.g. $\ba$;
bold upper case symbols for matrices and tensors, e.g. $\bA$;
the position within a tensor or vector is indicated with numeric subscripts, e.g. $\bA_{ij}$ with $i,j \in \mathbb{N}^+$;
calligraphic symbols for domains, e.g. $\mathcal{Q}$. 

\subsection{Intersection over Union (IoU)}
\label{sec:iou}
Given a pair of bounding box coordinates $(\bb_i, \bb_j)$, the \textit{Intersection over Union (IoU)}, also known as Jaccard index, is an evaluation metric used mainly in object detection tasks, which aims to evaluate how much the two bounding boxes refer to the same content in the image.
Specifically, it is defined as:
\begin{align}
    IoU(\bb_i, \bb_j) = \frac{|\bb_i \cap \bb_j|}{|\bb_i \cup \bb_j|},
\end{align}
where $|\bb_i \cap \bb_j|$ is the area of the box obtained by the intersection of  boxes $\bb_i$ and  $\bb_j$, while $|\bb_i \cup \bb_j|$ is the area of the box obtained by the union of boxes $\bb_i$ and  $\bb_j$.
It is invariant to the bounding boxes sizes, and it returns values that are strictly contained in the interval $[0, 1]\subset \mathbb{R}$, where $1$ means that the two bounding boxes refer to the same image area, while a score of $0$ means that the two bounding boxes do not overlap at all.
The fact that two bounding boxes that do not overlap have \textit{IoU} score equal to $0$, is the major issue of this metric: the zero value does not represent how much the two bounding boxes are far from each other.
For this reason, in its standard definition, the intersection over union  is mainly used as an evaluation metric rather than as a component of a loss function for learning.

\subsection{Complete Intersection over Union (CIoU)}
\label{sec:ciou}
In order to solve the issue of \textit{IoU} when considering it as a loss function, several alternative formulations were suggested in the literature, e.g.
\cite{rezatofighi2019generalized} proposed the \textit{Generalized IoU (GIoU)} loss, \cite{zheng2020distance} proposed the \textit{Distance IoU (DIoU)} loss, while only recently \cite{zheng2020enhancing} proposed the \textit{Complete IoU (CIoU)} loss, which has shown promising results and faster convergence than \textit{GIoU} and \textit{DIoU}.
It is defined as:
\begin{align}
\mathcal{L}_{CIoU}(\bb_i, \bb_j)
&=S\left(\bb_i, \bb_j\right)+D\left(\bb_i, \bb_j\right)+V\left(\bb_i, \bb_j\right)\\
S\left(\bb_i, \bb_j\right) &= 1 - IoU(\bb_i, \bb_j); \\
D\left(\bb_i, \bb_j\right) &= \frac{\rho\left(\boldsymbol{p_i}, \boldsymbol{p}_j\right)^{2}}{c^{2}}; \\
\label{eq:5}
V\left(\bb_i, \bb_j\right) &= \alpha \frac{4}{\pi^2} \left(\arctan\frac{wt_{j}}{ht_{j}} - \arctan\frac{wt_{i}}{ht_{i}}\right)
\end{align}
where $\bb_i$ and $\bb_j$ are two bounding boxes, $\boldsymbol{p_i}$ and $\boldsymbol{p}_j$ are their central points, $IoU(\bb_i, \bb_j)$ is the standard \textit{IoU}, $\rho$ is the euclidean distance between the given points, $c$ is the diagonal length of the \textit{convex hull} of the two bounding boxes, 
$\alpha$ is a trade-off parameter, $wt_i$ and $ht_i$ are the width and the height of the bounding box $\bb_i$, respectively.
Differently from the standard \textit{IoU}, the \textit{Complete IoU} is formulated in such a way to return meaningful values, leveraging the bounding boxes geometric shapes, even when two bounding boxes are not overlapped.

\section{Model}
As outlined in the main paper, our model follows a typical basic architecture for visual-textual grounding tasks. It is based on a two-stage approach in which, initially, a pre-trained object detector is used to extract, from a given image $\bI$, a set of $k$ bounding box proposals $\mathcal{P}_{\bI}=\{\bp_i\}_{i=1}^k$, where $\bp_i \in \mathbb{R}^4$, jointly with features \mbox{$H^v=\{\bh^v_i\}_{i=1}^k$,} where $\bh^v_i \in \mathbb{R}^{d}$, where $d$ is the number of returned features.
The features represent the internal object detector activation values before the classification layers and regression layer for bounding boxes.
Moreover, our model extracts the spatial features \mbox{$H^s=\{\bh^s_i\}_{i=1}^k$,} where $\bh^s_i \in \mathbb{R}^{5}$ from all the bounding boxes proposals. Specifically,  the spatial features for the proposal $\bp_i$ are defined as:
\begin{equation}
    \bh^s_i = \left[ \frac{x1}{wt}, \frac{y1}{ht}, \frac{x2}{wt}, \frac{y2}{ht}, \frac{(x2-x1) \times (y2-y1)}{wt \times ht}\right],
\end{equation}
where $(x1, y1)$ refers to the top-left bounding box corner, $(x2, y2)$ refers to the bottom-right bounding box corner, $wt$ and $ht$ are the  width and height of the image, respectively.  
%Moreover, our model extracts the spatial features \mbox{$H^s=\{\bh^s_i\}_{i=1}^k$,} where $\bh^s_i \in \mathbb{R}^{s}$ are the spatial features for the bounding box proposal $\bp_i$. 
We also assume that the object detector returns, for each  $\bp_i$, a probability distribution $Pr_{Cls}(\bp_i)$ over a set $Cls$ of predefined classes, i.e. the probability for each class $\xi\in Cls$ that the content of the bounding box $\bp_i$ belongs to $\xi$. %This information is typically returned by most of the object detectors, and it will be used to define our novel loss function.

Regarding the textual features extraction, given a noun phrase $\bq_j$, initially all its words $W^{\bq_j} = \{w^{\bq_j}_i\}_{i=1}^l$ are embedded in a set of vectors $E^{\bq_j} = \{\be^{\bq_j}_i\}_{i=1}^l$ where $\be^{\bq_j}_i \in \mathbb{R}^{w}$, where $w$ is the size of the embedding.
Then, our model applies a LSTM~\cite{hochreiter1997long} neural network to generate from the sequence of word embeddings only one new embedding  $\bh^\star_j$ for each phrase $\bq_j$.
This textual features extraction is defined as:
\begin{equation}
    \bh^\star_j = L1 \left(LSTM(E^{\bq_j}) \right),
\end{equation}
where $\bh^\star_j \in \mathbb{R}^t$ is the LSTM output of the last word in the noun phrase $\bq_j$, and $L1$ is the L1 normalization function.

Once vector  $\bh^\star_j$ has been generated from the noun phrase $\bq_j$, the model performs a multi-modal feature fusion operation in order to combine the information contained in $\bh^\star_j$ with each of the proposal bounding boxes $\bh^v_z$.
For this operation, we have decided to use a simple function that merges the multi-modal features together rather than relying on a more complex operator, such as bilinear-pooling or deep neural network architectures. %We leave the use of a more complex  fusion operator, that will lead to further improvements, for future work.
The multi-modal fusion component we adopted returns the set of new vectorial representations \mbox{$H^{||}=\{\bh^{||}_{jz}\}_{j \in [1,\ldots, m], z\in [1,\ldots, k]}$}, where  vectors $\bh^{||}_{jz}$ are defined as:
\begin{equation}
    \bh^{||}_{jz} = LR\left(\bW^{||} \left(\bh^\star_j \,||\, \bh^s_z \,||\, L1(\bh^v_z) \right) + \bb^{||} \right),
\end{equation}
where $||$ indicates the concatenation operator, \mbox{$\bh^{||}_{jz} \in \mathbb{R}^{c}$},  $LR$ indicates the leaky-relu activation function, \mbox{$\bW^{||} \in \mathbb{R}^{c\times (t+s+v)}$} is a matrix of weights, and $\bb^{||} \in \mathbb{R}^{c}$ is a bias vector.

Finally, the model predicts the probability $\bP_{jz}$ that a given noun phrase $\bq_j$ is referred to a proposal bounding box $\bp_z$ as:
\begin{equation}
    \bP_{jz} = 
    \frac{\exp(\bW^g \times \bh^{||}_{jz} + b^g)}
    {\sum{i=1}{k} \exp{(\bW^g \times \bh^{||}_{ji} + b^g)} },
\end{equation}
where $\bW^g \in \mathbb{R}^{1\times c}$ and $b^g \in \mathbb{R}$ are weights. 

Indeed, the representations $\bh^{||}_{jz}$ of the proposals bounding box features conditioned with the textual features 
can also be used to 
refine the proposal bounding box coordinates, that are generated by the object detector independently by
the textual features. Specifically, our model does not predicts new bounding box coordinates, but  offsets for the coordinates defined as 
\begin{equation}
    \bo_{jz} = \bW^\mathcal{B} \times \bh^{||}_{jz} + \bb^\mathcal{B},
\end{equation}
where $\bW^\mathcal{B} \in \mathbb{R}^{4\times c}$ and $\bb^\mathcal{B} \in \mathbb{R}^4$ are a matrix of weights and a bias vector, respectively. 
The final predicted bounding boxes coordinates are then obtained as the sum of the proposal bounding boxes coordinates with the predicted offsets.

\section{Datasets Details}
Flickr30k Entities and ReferIt constitute the two most common datasets used in the literature, although other datasets have been used (e.g., \cite{Chen_2020_CVPR, kazemzadeh2014referitgame, yu2016modeling, mao2016generation}).

The Flickr30k Entities dataset~\cite{plummer2015flickr30k, Flickr30k}  contains $32$K images, $275$K bounding boxes, $159$K sentences, and $360$K noun phrases. 
Each image is associated with five sentences with a variable number of noun phrases, and each noun phrase is associated with a set of bounding boxes ground truth coordinates.
Following all works in the literature, if a noun phrase corresponds to multiple ground truth bounding boxes, we merged the boxes and used their union region as its ground-truth.
On the contrary, a noun phrase with no associated bounding box was removed from the dataset.
We used the standard split for training, validation, and test set as defined in~\cite{plummer2015flickr30k}, consisting  of $30$K, $1$K, and $1$K images, respectively.

The ReferIt~\cite{KazemzadehOrdonezMattenBergEMNLP14} dataset contains $20$K images, $99$K bounding boxes, and $130$K noun phrases. 
This dataset differs from Flickr30k Entities since it does not contain sentences, which means that the noun phrases are mutually independent.
For this reason, the state-of-the-art models that depend on a sentence linking all the noun phrases, since they use a 
feature fusion operator that assumes the presence of the input sentence containing all the noun phrases, cannot be applied to it.
We used the same split as in~\cite{plummer2015flickr30k} that consists of $9$K images of training, $1$K images of validation, and $10$K images of test.

\section{Implementation Details}
\label{sec:implementation}
Our model extracts the words vocabulary using the SpaCy~\cite{honnibal2020spacy} framework for both  datasets.
Each word embedding is initialized using the GloVe~\cite{pennington2014glove} pre-trained weights, which our model does not train, while the remaining weights are initialized according to Xavier~\cite{glorot2010understanding}.
To compare objectively the experimental results with state-of-the-art models, we have used the same object detector adopted in~\cite{DBLP:conf/ijcai/YuYXZ0T18}, which consists of a Faster R-CNN pre-trained object detector~\cite{Anderson2017up-down} on the Visual Genome~\cite{krishna2017visual} dataset that uses ResNet-101 as backbone model\footnote{The ResNet-101 weights were pre-trained on COCO for initialization.}. The features associated to each bounding box are extracted from the  ResNet-101's layer \emph{pool5\_flat}.
Following~\cite{DBLP:conf/ijcai/YuYXZ0T18}, our object detector returns for each bounding box proposal a probability distribution over 1600 classes. 
We could have applied other object detectors or bounding box proposals which would have lead to further improvements, however this research direction is not related to the aim of this paper.
Our model adopts the normalized bounding boxes coordinates with the following representation:
\begin{equation}
    \bb = \left[ \frac{x1 + x2}{2}, \frac{y1 + y2}{2}, bwt, bht \right],
\end{equation}
where %$(x1, y1)$ refers to the top-left bounding box corner, $(x2, y2)$ refers to the bottom-right bounding box corner, 
$bwt$ and $bht$ are the  width and height of the bounding box, respectively.
%The spatial feature $\bh^s_i \in \mathbb{R}^{s}$ for the proposal $\bp_i$ are defined as:
%\begin{equation}
%    \bh^s_i = \left[ \frac{x1}{wt}, \frac{y1}{ht}, \frac{x2}{wt}, \frac{y2}{ht}, \frac{(x2-x1) \times (y2-y1)}{wt \times ht}\right],
%\end{equation}
%where $(x1, y1)$ refers to the top-left bounding box corner, $(x2, y2)$ refers to the bottom-right bounding box corner, $wt$ and $ht$ are the  width and height of the image, respectively.  

Regarding the parameter $alpha$ in Eq. \ref{eq:5}, we just used the value specified in \cite{zheng2020enhancing} which is identified by a specific formula. 

Regarding the application of our losses to the state-of-the-art DDPN~\cite{DBLP:conf/ijcai/YuYXZ0T18} model, we have used the authors' official code of the object detector to extract the bounding boxes proposals with their probabilities, and then we have re-implemented their DDPN model in Pytorch. 
Specifically, we implemented their model following the architecture and the hyper-parameters reported in their article, because the official implementation, as reported in the official repository\footnote{\url{https://github.com/XiangChenchao/DDPN}}, presents a slightly different architecture that leads to different results. 
On the re-implemented model, maintaining the same architecture and hyper-parameters, we have implemented our losses that lead to better results as reported in our main article.

% To apply our losses on the state-of-the-art DDPN~\cite{DBLP:conf/ijcai/YuYXZ0T18} model, we first extracted the bounding box proposals with their probabilities using the authors' official code and then re-implemented their model in Pytorch.
% Specifically, we implemented their model following the architecture and the hyper-parameter reported in their article, because the official implementation in their repository presents a slightly different architecture. 

\section{Qualitative Results}
\begin{figure*}
\begin{center}
\includegraphics[width=1\linewidth]{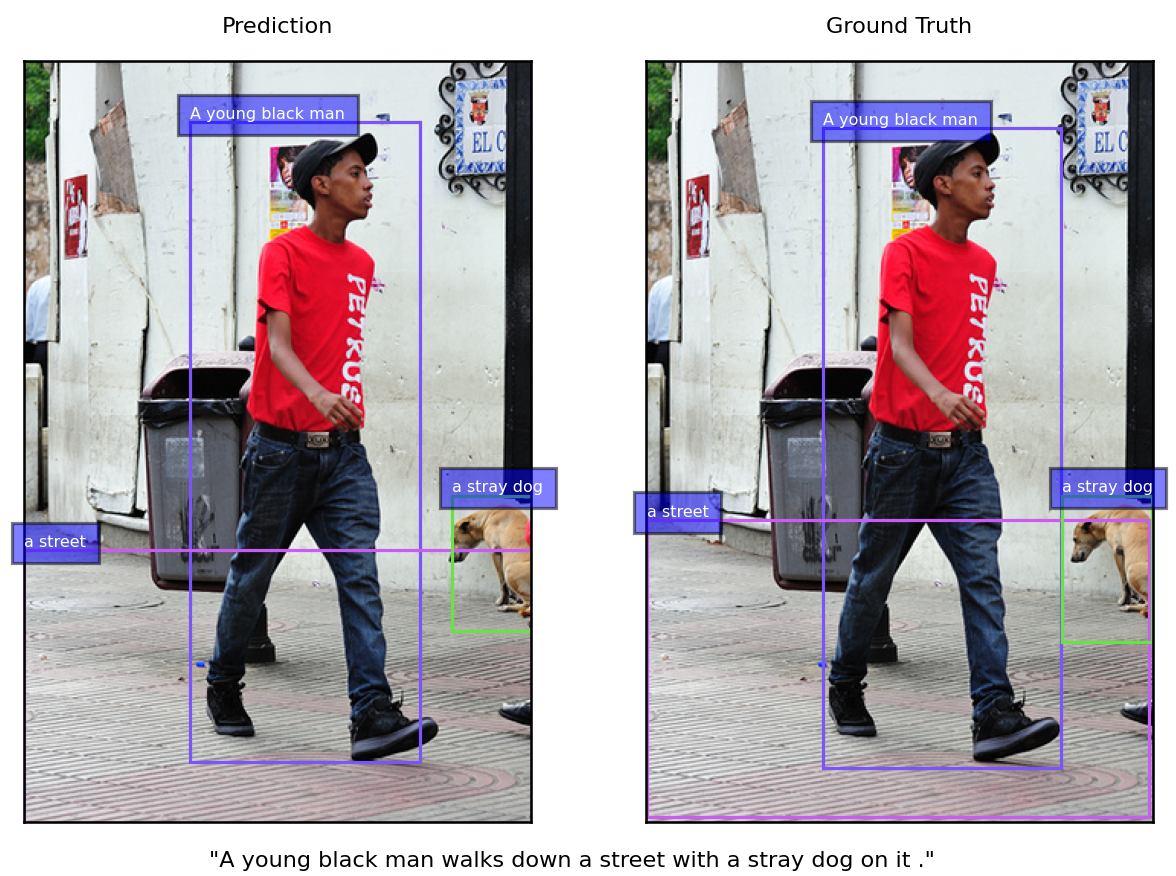}
\end{center}
\caption{\label{fig:flickr_1}Qualitative result obtained by our model on the Flickr30k Entities test set. All bounding boxes are predicted correctly.}
\end{figure*}

\begin{figure*}
\begin{center}
\includegraphics[width=1\linewidth]{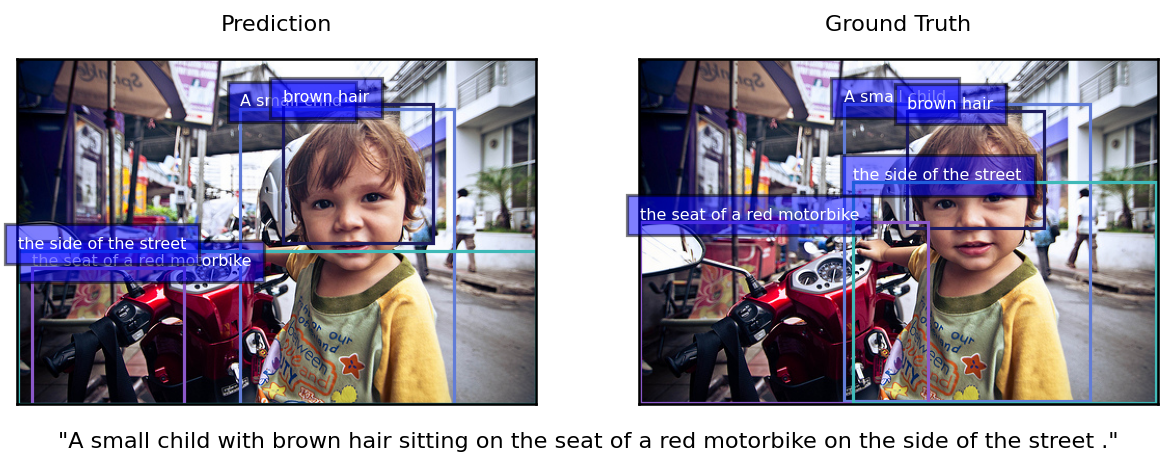}
\end{center}
\caption{\label{fig:flickr_2}Qualitative result obtained by our model on the Flickr30k Entities test set. The bounding boxes aligned with the queries ``the seat of a red motorbike" and ``the side of the street" present an intersection over union value with their ground truths that is lower than 0.5.}
\end{figure*}

\begin{figure*}
\begin{center}
\includegraphics[width=1\linewidth]{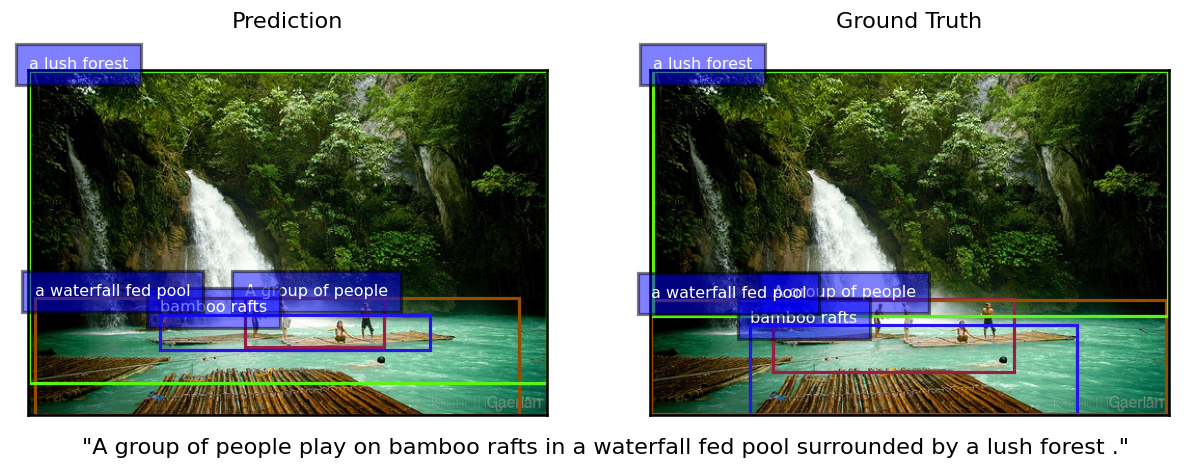}
\end{center}
\caption{\label{fig:flickr_3}Qualitative result obtained by our model on the Flickr30k Entities test set. The bounding boxes aligned with the queries ``A group of people" and ``bamboo rafts" present an intersection over union value with their ground truths that are lower than 0.5.}
\end{figure*}

\begin{figure*}
\begin{center}
\includegraphics[width=1\linewidth]{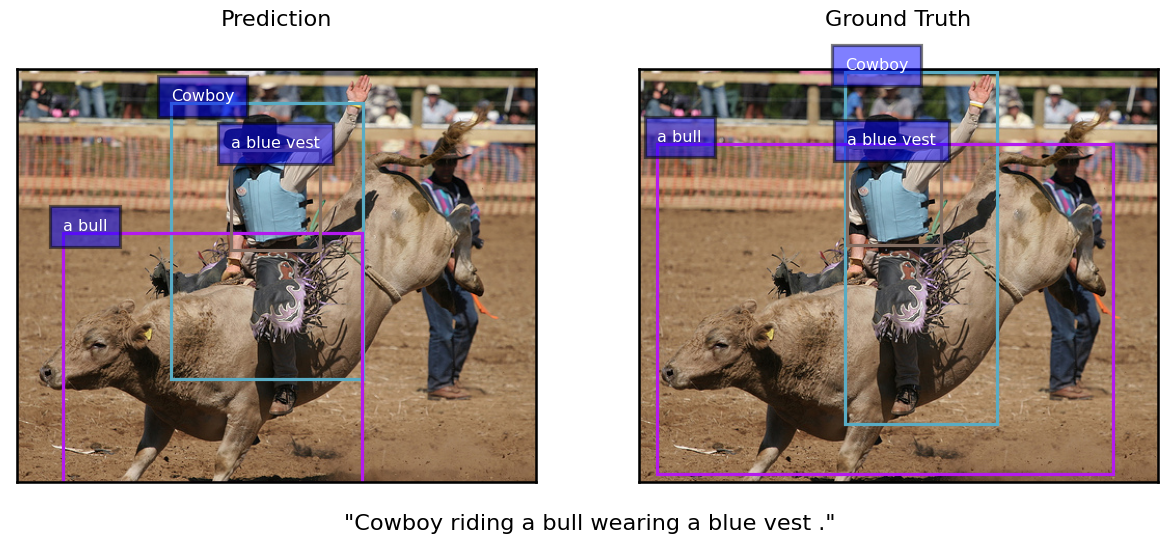}
\end{center}
\caption{\label{fig:flickr_4}Qualitative result obtained by our model on the Flickr30k Entities test set. The bounding box aligned with the query ``a bull" presents an intersection over union value with its ground truth that is lower than 0.5.}
\end{figure*}

\begin{figure*}
\begin{center}
\includegraphics[width=1\linewidth]{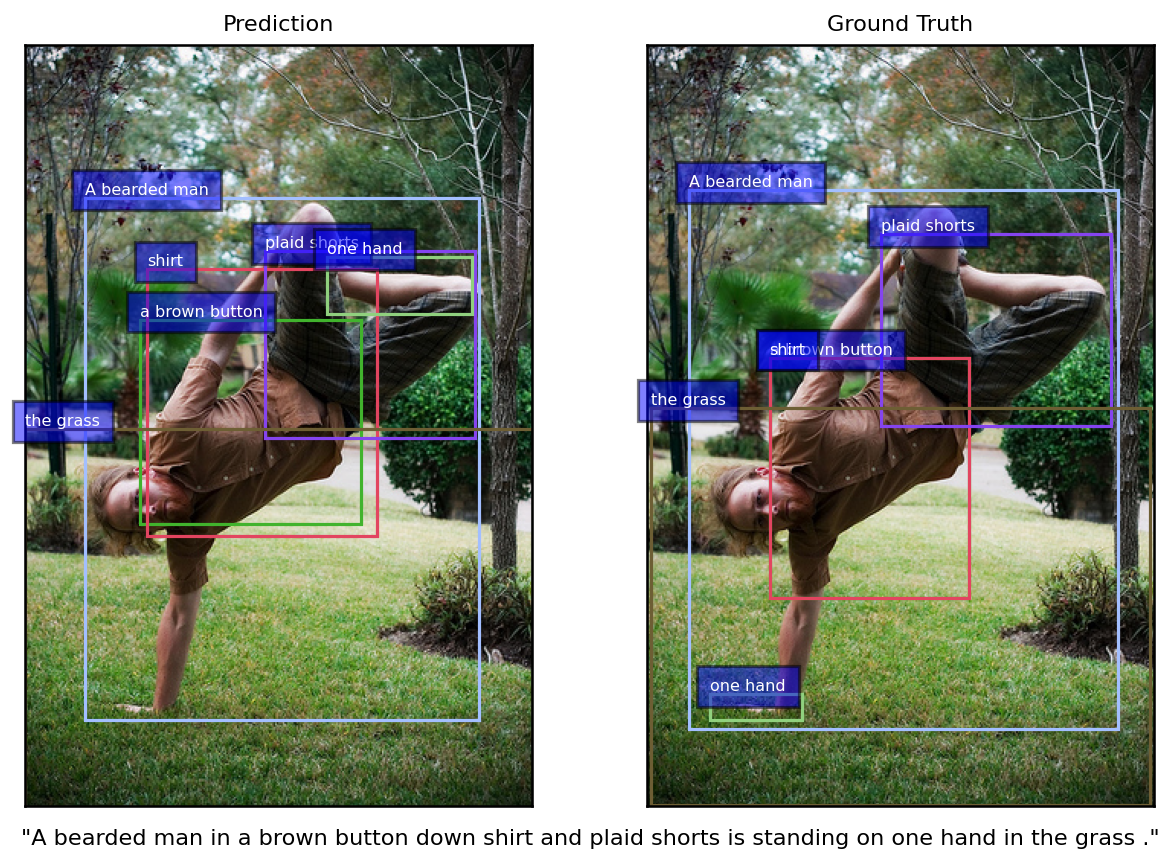}
\end{center}
\caption{\label{fig:flickr_5}Qualitative result obtained by our model on the Flickr30k Entities test set. The bounding boxes aligned with the queries ``shirt" and ``one hand" present an intersection over union value with their ground truths that are lower than 0.5.}
\end{figure*}

\begin{figure*}
\begin{center}
\includegraphics[width=1\linewidth]{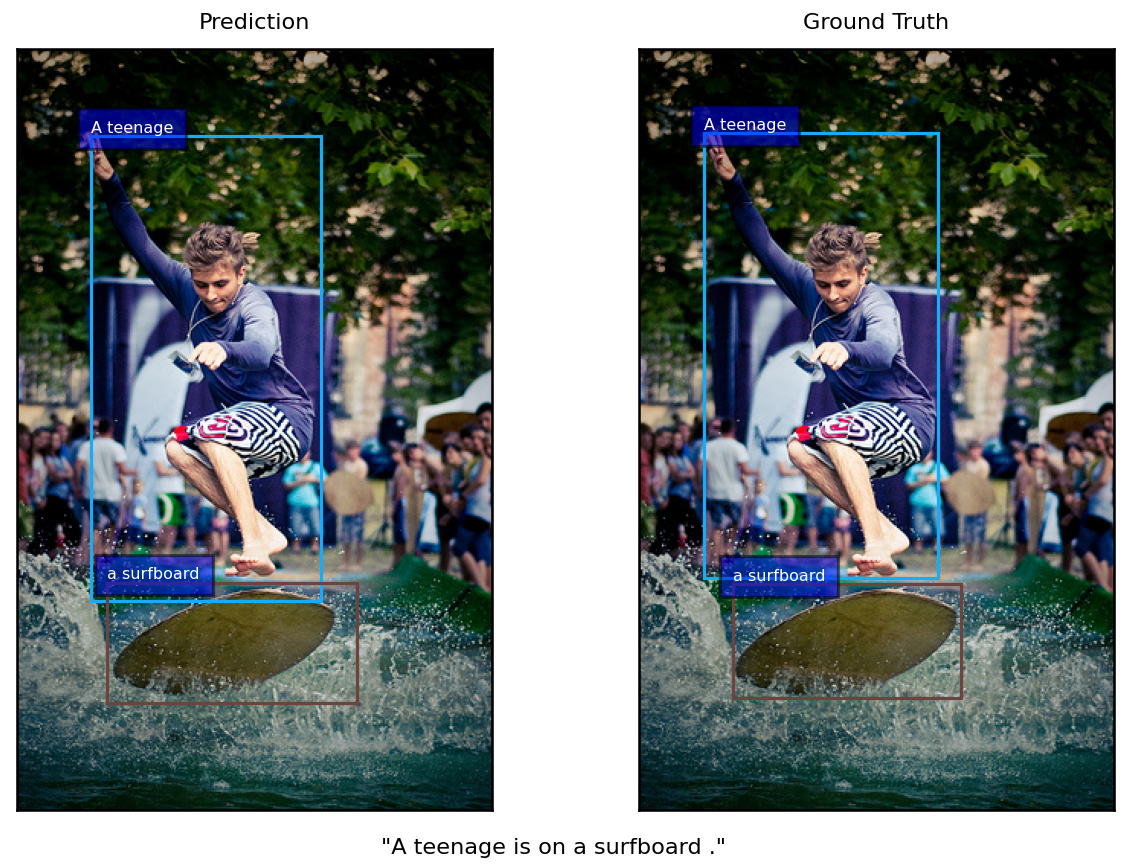}
\end{center}
\caption{\label{fig:flickr_6}Qualitative result obtained by our model on the Flickr30k Entities test set. All bounding boxes are predicted correctly.}
\end{figure*}

\begin{figure*}
\begin{center}
\includegraphics[width=1\linewidth]{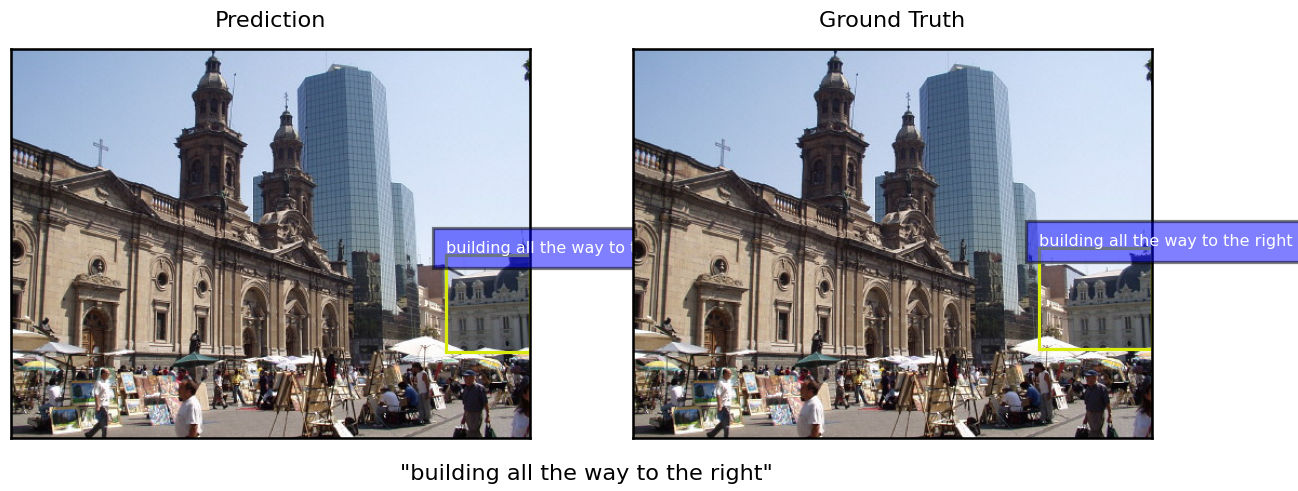}
\end{center}
\caption{\label{fig:referit_1}Qualitative result obtained by our model on the ReferIt test set. The bounding box is predicted correctly.}
\end{figure*}

\begin{figure*}
\begin{center}
\includegraphics[width=1\linewidth]{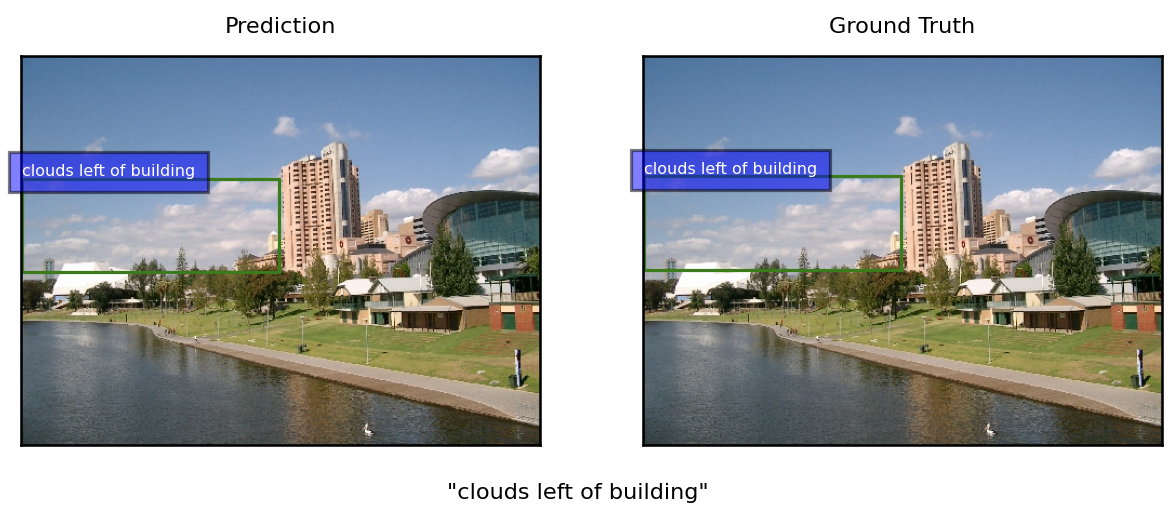}
\end{center}
\caption{\label{fig:referit_2}Qualitative result obtained by our model on the ReferIt test set. The bounding box is predicted correctly.}
\end{figure*}

\begin{figure*}
\begin{center}
\includegraphics[width=1\linewidth]{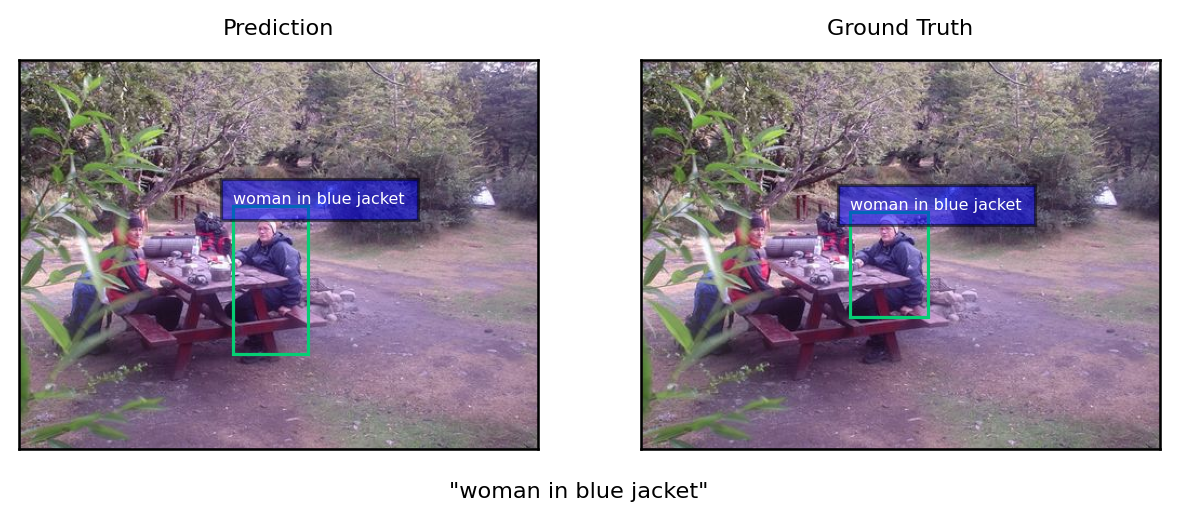}
\end{center}
\caption{\label{fig:referit_3}Qualitative result obtained by our model on the ReferIt test set. The bounding box is predicted correctly.}
\end{figure*}

\begin{figure*}
\begin{center}
\includegraphics[width=1\linewidth]{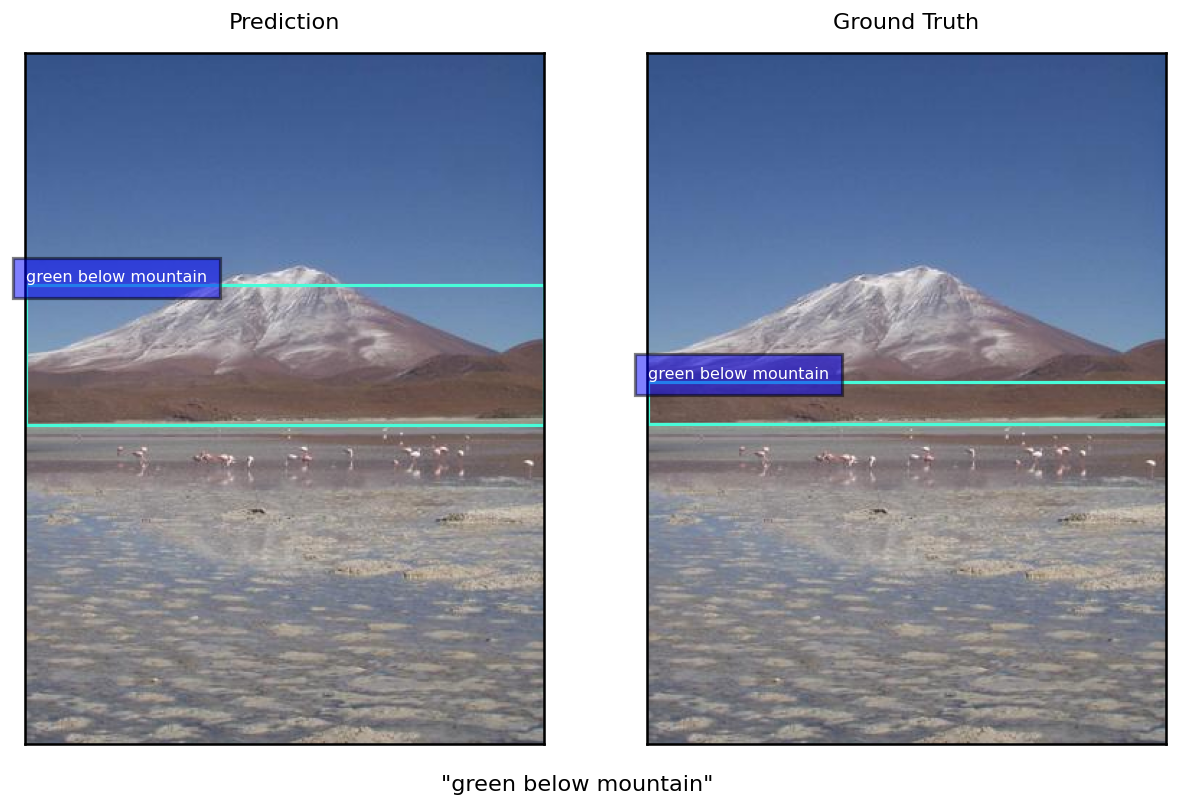}
\end{center}
\caption{\label{fig:referit_4}Qualitative result obtained by our model on the ReferIt test set. The predicted bounding box presents an intersection over union value with the ground truth of 0.30.}
\end{figure*}

\begin{figure*}
\begin{center}
\includegraphics[width=1\linewidth]{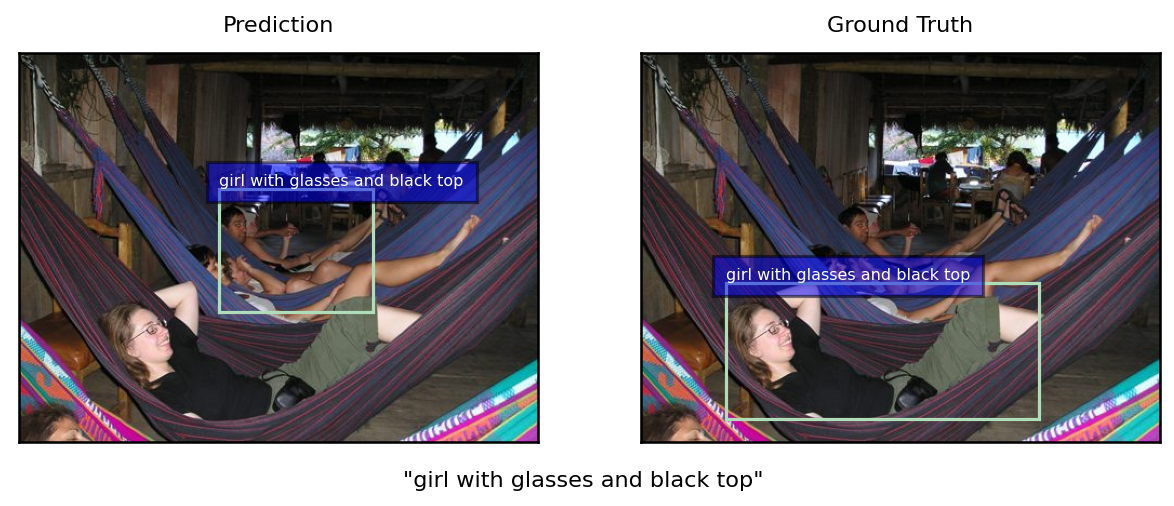}
\end{center}
\caption{\label{fig:referit_5}Qualitative result obtained by our model on the ReferIt test set. The predicted bounding box presents an intersection over union value with the ground truth of 0.08.}
\end{figure*}

\begin{figure*}
\begin{center}
\includegraphics[width=1\linewidth]{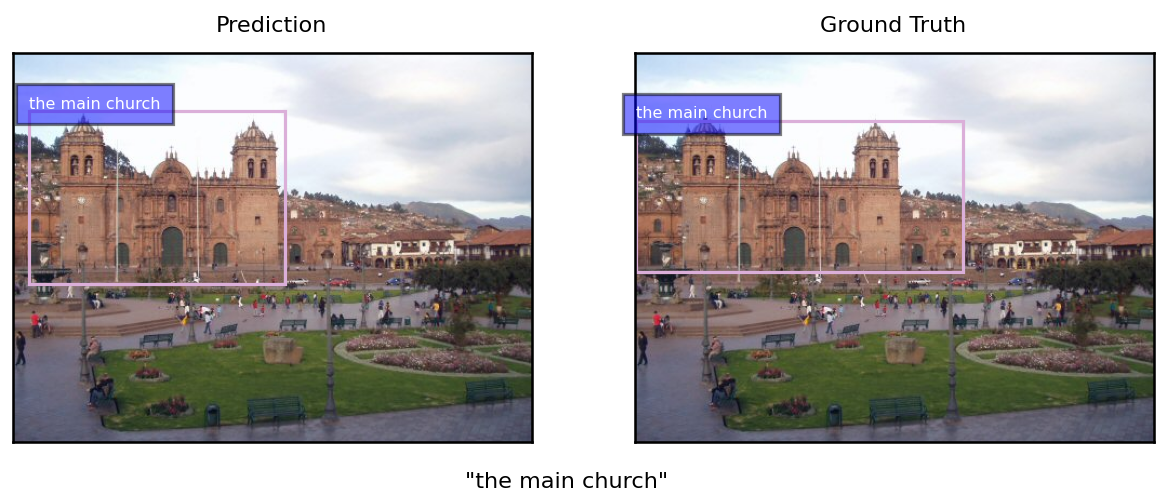}
\end{center}
\caption{\label{fig:referit_6}Qualitative result obtained by our model on the ReferIt test set. The bounding box is predicted correctly.}
\end{figure*}

In Figures~\ref{fig:flickr_1}-\ref{fig:referit_6}, we have reported some qualitative results obtained by our model in both Flickr30k Entities and ReferIt datasets. 
Figures \ref{fig:flickr_1}, \ref{fig:flickr_2}, \ref{fig:flickr_3}, \ref{fig:flickr_4}, \ref{fig:flickr_5}, \ref{fig:flickr_6} are examples of the Flickr30k Entities test set images, while Figures \ref{fig:referit_1}, \ref{fig:referit_2}, \ref{fig:referit_3}, \ref{fig:referit_4}, \ref{fig:referit_5}, \ref{fig:referit_6} are examples of the ReferIt test set. We can see that in both the datasets, very often the predicted bounding boxes that have an intersection over union value under 0.5, are still close to the ground truths bounding boxes. Only in Figure  \ref{fig:flickr_5}, the model predicts a bounding box for the query ``one hand" that is located very far from its ground truth.



%%%%%%%%% REFERENCES
\bibliographystyle{ACM-Reference-Format}
\bibliography{bibliography}